\documentclass[11pt]{article}

\pdfoutput=1 %

\usepackage[utf8]{inputenc}
\usepackage{graphicx}
\usepackage{color,colortbl}
\usepackage[svgnames]{xcolor}
\usepackage{booktabs}
\usepackage{enumitem}
\usepackage{array}
\usepackage{url}
\usepackage[hidelinks,colorlinks=true,allcolors=MediumBlue]{hyperref}
\usepackage[authoryear,sort&compress,square]{natbib}
\usepackage{amsmath}
\usepackage{makecell}
\usepackage{amssymb}
\usepackage{multirow}
\usepackage{microtype} %
\usepackage{abstract}
\usepackage[export]{adjustbox}
\usepackage{layouts}
\usepackage{fontawesome}

\usepackage[bottom=30pt,top=80pt,left=124pt,bottom=80pt,right=124pt,headsep=0pt,footskip=30pt]{geometry}

\usepackage[scaled]{helvet}
\usepackage[T1]{fontenc}
\usepackage{sectsty}
\allsectionsfont{\sffamily}

\makeatletter
\def\@maketitle{%
  \newpage
  \null
  \vspace{-2em}
  \sffamily
  \begin{center}%
  \let \footnote \thanks
    {\Huge\bfseries\@title \par}%
    \vskip 2.4em%
    {\large \@author}
    \vskip 1em%
  \end{center}%
  \par
  \vskip 1.5em}
\makeatother

\makeatletter
\pretocmd{\NAT@citex}{%
  \let\NAT@hyper@\NAT@hyper@citex
  \def\NAT@postnote{#2}%
  \setcounter{NAT@total@cites}{0}%
  \setcounter{NAT@count@cites}{0}%
  \forcsvlist{\stepcounter{NAT@total@cites}\@gobble}{#3}}{}{}
\newcounter{NAT@total@cites}
\newcounter{NAT@count@cites}
\def\NAT@postnote{}

\def\NAT@hyper@citex#1{%
  \stepcounter{NAT@count@cites}%
  \hyper@natlinkstart{\@citeb\@extra@b@citeb}#1%
  \ifnumequal{\value{NAT@count@cites}}{\value{NAT@total@cites}}
    {\ifNAT@swa\else\if*\NAT@postnote*\else%
     \NAT@cmt\NAT@postnote\global\def\NAT@postnote{}\fi\fi}{}%
  \ifNAT@swa\else\if\relax\NAT@date\relax
  \else\NAT@@close\global\let\NAT@nm\@empty\fi\fi%
  \hyper@natlinkend}
\renewcommand\hyper@natlinkbreak[2]{#1}

\patchcmd{\NAT@citex}
  {\ifNAT@swa\else\if*#2*\else\NAT@cmt#2\fi
   \if\relax\NAT@date\relax\else\NAT@@close\fi\fi}{}{}{}
\patchcmd{\NAT@citex}
  {\if\relax\NAT@date\relax\NAT@def@citea\else\NAT@def@citea@close\fi}
  {\if\relax\NAT@date\relax\NAT@def@citea\else\NAT@def@citea@space\fi}{}{}

\makeatother

\let\cite\citep

\newcommand{\ourmodel}[1]{AFM#1}
\newcommand{\committee}{iTeC}
\newcommand{\committeefull}{Iterative teaching committee}
\newcommand{\mistral}{Mistral-7B} %
\newcommand{\dbrx}{DBRX-Instruct} %

\newcommand{\phimini}{Phi-3-mini} %
\newcommand{\gemmal}{Gemma-7B} %
\newcommand{\gpts}{GPT3.5} %
\newcommand{\gptl}{GPT4} %

\newcommand{\llamathrees}{Llama-3-8B}

\newcommand{\inputhide}[1]{} %

\newcommand{\bftab}{\fontseries{b}\selectfont}

\definecolor{light}{rgb}{0.5, 0.5, 0.5}

\title{Apple Intelligence Foundation Language Models}
\author{Apple}
\date{July 2024}

\hypersetup{pdfauthor={Apple},pdftitle={Apple Intelligence Foundation Language Models}}

\begin{document}

\maketitle

\begin{abstract}
\sffamily
\noindent We present foundation language models developed to power Apple Intelligence features, including a $\sim$3~billion parameter model designed to run efficiently on devices and a large server-based language model designed for Private Cloud Compute~\cite{ApplePCC}.
These models are designed to perform a wide range of tasks efficiently, accurately, and responsibly.
This report describes the model architecture, the data used to train the model, the training process, how the models are optimized for inference, and the evaluation results. We highlight our focus on Responsible AI and how the principles are applied throughout the model development.
\end{abstract}

\section{Introduction}

At the 2024 Worldwide Developers Conference, we introduced Apple Intelligence, a personal intelligence system integrated deeply into iOS 18, iPadOS 18, and macOS Sequoia.

Apple Intelligence consists of multiple highly-capable generative models that are fast, efficient, specialized for our users' everyday tasks, and can adapt on the fly for their current activity. The foundation models built into Apple Intelligence have been fine-tuned for user experiences such as writing and refining text, prioritizing and summarizing notifications, creating playful images for conversations with family and friends, and taking in-app actions to simplify interactions across apps.

\begin{figure}[!ht]
    \includegraphics[width=1.2\textwidth,alt={Modeling Overview Diagram.},center]{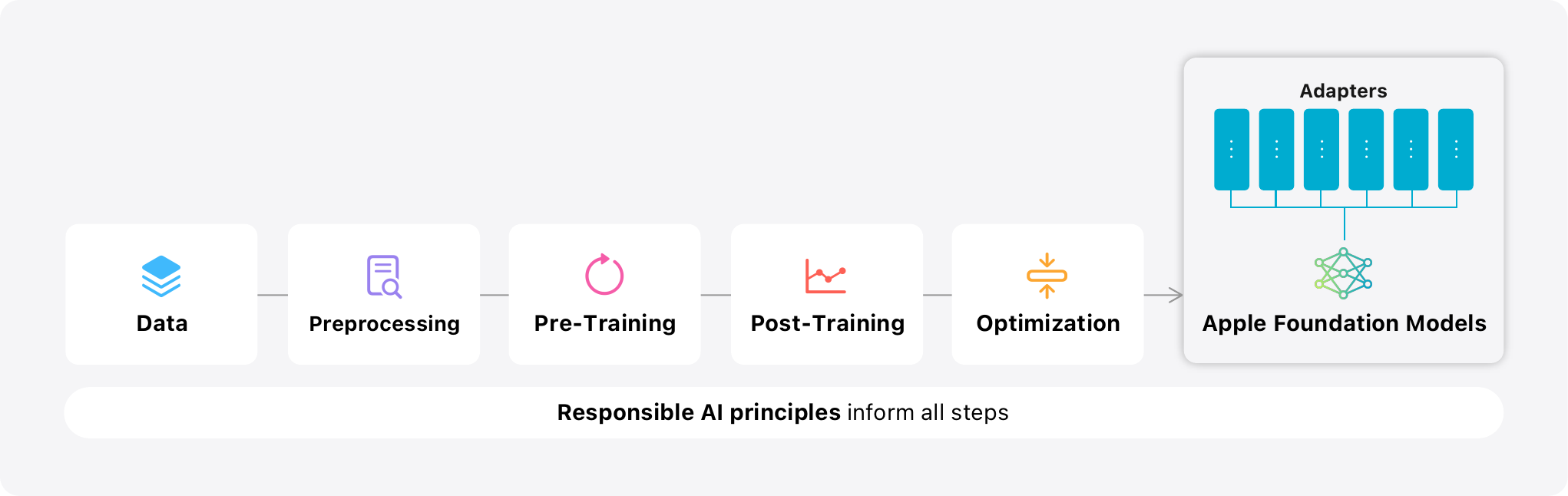}
    \caption{Modeling overview for the Apple foundation models.}
    \label{fig:overview}
\end{figure}

In this report we will detail how two of these models---\ourmodel{-on-device} (AFM stands for \emph{Apple Foundation Model}), a $\sim$3~billion parameter language model, and \ourmodel{-server}, a larger server-based language model---have been built and adapted to perform specialized tasks efficiently, accurately, and responsibly (\autoref{fig:overview}). These two foundation models are part of a larger family of generative models created by Apple to support users and developers; this includes a coding model (based on an \ourmodel{} language model) to build intelligence into Xcode, as well as a diffusion model to help users express themselves visually, for example, in the Messages app.

Apple Intelligence is designed with Apple's core values at every step and built on a foundation of industry-lead privacy protection. Additionally, we have created Responsible AI principles to guide how we develop AI tools, as well as the models that underpin them:

\begin{enumerate}
    \item \textbf{Empower users with intelligent tools}: We identify areas where AI can be used responsibly to create tools for addressing specific user needs. We respect how our users choose to use these tools to accomplish their goals.
    \item \textbf{Represent our users}: We build deeply personal products with the goal of representing users around the globe authentically. We work continuously to avoid perpetuating stereotypes and systemic biases across our AI tools and models.
    \item \textbf{Design with care}: We take precautions at every stage of our process, including design, model training, feature development, and quality evaluation to identify how our AI tools may be misused or lead to potential harm. We will continuously and proactively improve our AI tools with the help of user feedback.
    \item \textbf{Protect privacy}: We protect our users' privacy with powerful on-device processing and groundbreaking infrastructure like Private Cloud Compute. We do not use our users' private personal data or user interactions when training our foundation models.
\end{enumerate}

These principles are reflected at every stage of the architecture that enables Apple Intelligence and connects features and tools with specialized models.

In the remainder of this report, we provide details on decisions such as: how we develop models that are highly capable, fast, and power-efficient; how we approach training these models; how our adapters are fine-tuned for specific user needs; and how we evaluate model performance for both helpfulness and unintended harm.

\section{Architecture}
\label{sec:architecture}
The \ourmodel{} base models are dense decoder-only models that build on the Transformer architecture~\cite{NIPS2017_3f5ee243}, with the following design choices:

\begin{itemize}[itemsep=0pt]
    \item A shared input/output embedding matrix~\cite{Press2016UsingTO} to reduce memory usage for parameters.
    \item Pre-Normalization~\cite{nguyen-salazar-2019-transformers} with RMSNorm~\cite{rmsnorm} for training stability.
    \item Query/key normalization~\cite{wortsman2023} to improve training stability.
    \item Grouped-query attention (GQA)~\cite{ainslie2023gqa} with 8 key-value heads to reduce the KV-cache memory footprint.
    \item The {SwiGLU} activation~\cite{shazeer2020gluvariantsimprovetransformer} for higher efficiency.
    \item RoPE~\cite{su2024roformer} positional embeddings with the base frequency set to $500\mathrm{k}$ for long-context support.
\end{itemize}

\autoref{tab:architecture} provides some details about \ourmodel{-on-device} dimensions.

\begin{table}[h!]
\centering
\begin{tabular}{@{}lr@{}}
    \toprule
    Model dimension & $3072$ \\
    Head dimension & $128$ \\
    Num query heads & $24$ \\
    Num key/value heads & $8$ \\
    Num layers & $26$ \\
    \midrule
    Num non-embedding params (B) & $2.58$ \\
    Num embedding params (B) & $0.15$ \\
    \bottomrule
\end{tabular}
\caption{\ourmodel{-on-device} dimensions.}
\label{tab:architecture}
\end{table}

\section{Pre-training} %

Our AFM pre-training process plays a critical role in developing highly capable language models to power a host of Apple Intelligence features that can help and support users. We focus on efficiency and data quality at every step in order to pre-train for a high-quality end-to-end user experience with efficient and low-latency models. 

\subsection{Data}
The AFM pre-training dataset consists of a diverse and high quality data mixture.
This includes data we have licensed from publishers, curated publicly-available or open-sourced datasets, and publicly available information crawled by our web-crawler, Applebot~\cite{AppleBot}. We respect the right of webpages to opt out of being crawled by Applebot, using standard robots.txt directives. 

Given our focus on protecting user privacy, we note that no private Apple user data is included in the data mixture. Additionally, extensive efforts have been made to exclude profanity, unsafe material, and personally identifiable information from publicly available data (see \autoref{sSafety} for more details). Rigorous decontamination is also performed against many common evaluation benchmarks. 

We find that data quality, much more so than quantity, is the key determining factor of downstream model performance. In the following, we provide more details about key components of the data mixture.

\subsubsection{Web pages}
\label{sec:web-pages}
We crawl publicly available information using our web crawler, Applebot~\cite{AppleBot}, and respect the rights of web publishers to opt out of Applebot using standard robots.txt directives.
Plus, we take steps to exclude pages containing profanity and apply filters to remove certain categories of personally identifiable information (PII).
The remaining documents are then processed by a pipeline which performs quality filtering and plain-text extraction, more specifically:

\begin{enumerate}[itemsep=0pt]
    \item Body extraction is performed using a combination of Safari's reader mode and the Boilerpipe~\cite{boilerpipe19} algorithm.
    \item Safety and profanity filtering, using heuristics and model-based classifiers.
    \item Global fuzzy de-duplication using locality-sensitive n-gram hashing.
    \item Extensive quality filtering using heuristics and model-based classifiers~\cite{kong2024,dclmshankar}.
    \item Decontamination against 811 common pre-training benchmarks, filtering entire documents upon 4-13 gram collisions with any of the benchmark datasets, unless the collision-count for a given n-gram reaches a ``common-usage" threshold of 1000.
\end{enumerate}

\subsubsection{Licensed datasets}\label{sec:licensed_data}
We go to lengths to identify and license a limited amount of high-quality data from publishers.
These licensed datasets provide a natural source of diverse and high quality long-context data, so we include them as part of the data mixture for the continued and context-lengthening stages of pre-training (see \autoref{sec:continued_pretraining} and \ref{sec:context_lengthening} for more details).
We decontaminate sections of publisher licensed data the same way we decontaminate web pages (\autoref{sec:web-pages}). 

\subsubsection{Code}
\label{sec:pre_train_code}
Code data is obtained from license-filtered\footnote{
Using MIT, Apache, BSD, CC0, CC-BY, Unlicensed, ISC, and Artistic Licenses.
} open source repositories on GitHub. The bulk of the code data covers 14 common programming languages, including: Swift, Python, C, Objective-C, C++, JavaScript, Java, and Go.
The data is de-duplicated, further filtered for PII and quality, and decontaminated in the same fashion as in \autoref{sec:web-pages}.

\subsubsection{Math}
We integrate two categories of high-quality data sourced from the web. The first category is a Math Q\&A dataset, comprising 3 billion tokens from 20 web domains rich in math content. We extract the questions and answers by identifying relevant tags from HTML pages. The second category is a collection of 14 billion tokens from web pages such as math forums, blogs, tutorials, and seminars. To filter these web pages, we used a specialized pipeline that includes a math tag filter with a collection of 40 strings to identify mathematical templates, a math symbol filter with a collection of 350 Unicode and LaTeX symbols to identify math content, a quality filter powered by a language model classifier specifically designed for math~\cite{kong2024}, and a domain filter that processes all web pages from domains manually labeled by humans. We applied these filters, followed by deduplication, decontamination, and PII removal to produce the final dataset.

\subsubsection{Public datasets}
We evaluated and selected a number of high-quality publicly-available datasets with licenses that permit use for training language models. 
Then, we filtered the datasets to remove personally identifiable information before including them in the pre-training mixture.

\subsubsection{Tokenizer}
We use a \emph{byte-pair encoding}~(BPE) tokenizer, following the implementation from SentencePiece. All numbers are split into individual digits and we use byte-fallback to decompose unknown UTF-8 characters into byte tokens. We do not enable Unicode normalization.
The total vocabulary size is 100k and 49k tokens for \ourmodel{-server} and \ourmodel{-on-device}, respectively.

\subsection{Recipe}
We break \ourmodel{} pre-training into three distinct stages: 1. \emph{core} which consumes most of the compute budget, 2. \emph{continued}, where we down-weight the lower-quality bulk web-crawl data, favoring a higher code and math weight instead combined with inclusion of the licensed data described in \autoref{sec:licensed_data}, 3. \emph{context-lengthening} which is similar to another \emph{continued} pre-training stage, but conducted at longer sequence length and with synthetic long-context data included in the mixture.

Details about model quality after each of the three pre-training stages (alongside additional metrics for \ourmodel{} derived from our internal benchmark implementations) are in \autoref{sec:pre-train_stage_breakdown_eval}, and \autoref{sec:ruler_long_context} examines \ourmodel{-server}'s long-context capabilities.

All three stages use decoupled weight decay~\cite{decoupledweightdecay} for regularization, as well as a simplified version of $\mu$Param~\cite{yang2022tensorprogramsvtuning}, similar to what is described as $\mu$Param (simple) in~\cite{wortsman2023}. Thus far we have not found more sophisticated parameter norm controls to be necessary at these scales. All stages maintain sharded model and optimizer states in \texttt{float32}, casting to \texttt{bfloat16} for the forward and backward passes for efficiency.

\subsubsection{Core pre-training} \label{sec:core_pretraining}
\ourmodel{-server} core training is conducted from scratch, while \ourmodel{-on-device} is distilled and pruned from a larger model. 

\paragraph{\textbf{\ourmodel{-server}}:}
We train \ourmodel{-server} from scratch for 6.3T tokens on 8192 TPUv4 chips, using a sequence length of 4096 and a batch-size of 4096 sequences. The batch size was determined using a scaling law fit to model size and compute budget, however we find that downstream results are relatively insensitive to a fairly wide range of batch sizes, and expect that any value between $0.5\times$ and $2\times$ the predicted batch size would have yielded similar results (the predicted optimum was in fact $\sim$3072, but 4096 allowed for better chip utilization).
We perform a learning rate sweep using a proxy model with a model dimension of 768, finding that the optimum learning rate spans 0.01-0.02, so we choose 0.01 to be conservative. Linear layers will have an effective learning rate scaled by $\sim$0.1 due to the use of $\mu$Param (simple).\footnote{In scaling law experiments we find that $\mu$Param (simple) stabilizes the optimal learning rate as model size increases, although extrapolating to very significantly deeper and/or larger models does exhibit a slight left-shift beyond what is accounted for.}
 
We use a tuned decoupled weight decay of $3.16\mathrm{e}{-4}$, finding that it works well across all tested model sizes and compute budgets. The learning rate schedule includes a linear warm-up for 5000 steps, followed by cosine decay to 0.005 of the peak over the remainder of training. For further details on the optimizer, see \autoref{sec:pretrainoptimizer}. \autoref{sec:core_pretrain_recipe_ablation} compares the \ourmodel{} core pre-training recipe to a more typical configuration.

\paragraph{\textbf{\ourmodel{-on-device}}:}
For the on-device model, we found that knowledge distillation~\cite{hinton2015distilling} and structural pruning are effective ways to improve model performance and training efficiency. These two methods are complementary to each other and work in different ways. More specifically, before training \ourmodel{-on-device}, we initialize it from a pruned 6.4B model (trained from scratch using the same recipe as \ourmodel{-server}), using pruning masks that are learned through a method similar to what is described in~\cite{wang2020structured,xiasheared}. The key differences are: (1) we only prune the hidden dimension in the feed-forward layers; (2) we use Soft-Top-K masking~\cite{lei2023conditional} instead of HardConcrete masking~\cite{louizos2018learning}; (3) we employ the same pre-training data mixture as the core phase to learn the mask, training for 188B tokens. Then, during the core pre-training of \ourmodel{-on-device}, a distillation loss is used by replacing the target labels with a convex combination of the true labels and the teacher model's top-1 predictions, (with 0.9 weight assigned to the teacher's labels), training for a full 6.3T tokens.  We observe that initializing from a pruned model improves both data efficiency and the final benchmark results by 0-2\%, whilst adding distillation boosts MMLU and GSM8K by about 5\% and 3\% respectively. More detailed ablation results can be found in \autoref{sec:abl_prune_distill}. All training hyper-parameters except for batch-size are kept the same as \ourmodel{-server}.

\subsubsection{Continued pre-training}\label{sec:continued_pretraining}
For both models we perform continued pre-training at a sequence length of 8192, with another 1T tokens from a mixture that upweights math and code, and down-weights the bulk web-crawl. We also include the licensed data described in \autoref{sec:licensed_data}. We use a peak learning rate of $3\mathrm{e}{-4}$ and decoupled weight decay of $1\mathrm{e}{-5}$, and 1000 warm-up steps with a final learning rate decay to 0.001 of peak, differently to core pre-training. Other settings (batch size, etc) are carried over. We did not find a distillation loss to be helpful here for \ourmodel{-on-device}, unlike in core pre-training, so the recipe is identical to that used for \ourmodel{-server}.

\subsubsection{Context lengthening}\label{sec:context_lengthening}
Finally, we conduct a further 100B tokens of continued pre-training at a sequence length of $32768$ tokens, using the data mixture from the continued pre-training stage, augmented with synthetic long-context Q\&A data. We also increase the RoPE base frequency from 500k to $6315089$, following the scaling laws described in~\cite{liu2024scalinglawsropebasedextrapolation}, with the expectation that this will allow for better short-to-long generalization---which is desirable given that the majority of our pre-training data is comprised of documents that are significantly shorter than 32k tokens long. The recipe is similar to that used for continued pre-training. We examine the long-context performance of \ourmodel{-server} in \autoref{sec:ruler_long_context}.

\subsubsection{Optimizer}
\label{sec:pretrainoptimizer}
We choose to use a variant of RMSProp~\cite{tieleman2012lecture} with momentum for \ourmodel{} pre-training. In particular, we divide the raw gradient by the square-root of a bias-corrected exponential moving average of the squared gradient to produce an instantaneous update, which is clipped to a maximum norm of $1.0$ per parameter block, before then further smoothing this estimate over steps with an exponential moving average without bias-correction to produce the net update. Unless otherwise noted, the smoothing constants for both the squared gradient ($\beta_2$) and the update ($\beta_1$) are set to $0.95$. A small constant $\epsilon=1\mathrm{e}{-30}$ is added to the instantaneous squared gradient prior to smoothing, for numerical stability.

\paragraph{}The smoothed updates are scaled by the learning rate, weight-decay is added, and then scheduled decay is applied to form the final weight delta. As an additional guard for stability, prior to the optimizer we clip the global gradient norm to $1.0$. For a recipe ablation against a more typical configuration, see \autoref{sec:core_pretrain_recipe_ablation}.

\subsection{Training infrastructure}
The \ourmodel{} models are pre-trained on v4 and v5p Cloud TPU clusters with the AXLearn framework~\cite{axlearn}, a JAX~\cite{jax2018github} based deep learning library designed for the public cloud.
Training is conducted using a combination of tensor, fully-sharded-data-parallel, and sequence parallelism, allowing training to scale to a large number of model parameters and sequence lengths at high utilization.
This system allows us to train the \ourmodel{} models efficiently and scalably, including \ourmodel{-on-device}, \ourmodel{-server}, and larger models.

\paragraph{}\ourmodel{-server} was trained on 8192 TPUv4 chips provisioned as $8 \times 1024$ chip slices, where slices are connected together by the data-center network (DCN)~\cite{palm}. Only data-parallelism crosses the slice boundary, other types of state sharding are within-slice only as the within-slice interconnect bandwidth is orders of magnitude higher than the DCN. The sustained model-flop-utilization (MFU) for this training run was approximately 52\%. \ourmodel{-on-device} was trained on one slice of 2048 TPUv5p chips.

\section{Post-Training}
While Apple Intelligence features are powered through adapters on top of the base model (see \autoref{sec:features} for a deep-dive on the adapter architecture), empirically we found that improving the general-purpose post-training lifts the performance of all features, as the models have stronger capabilities on instruction following, reasoning, and writing.

We conduct extensive research in post-training methods to instill general-purpose instruction following and conversation capabilities to the pre-trained \ourmodel{} models.
Our goal is to ensure these model capabilities are aligned with Apple's core values and principles, including our commitment to protecting user privacy, and our Responsible AI principles.
Our post-training efforts include a series of data collection and generation, instruction tuning, and alignment innovations.
Our post-training process contains two stages: supervised fine-tuning (SFT) and reinforcement learning from human feedback (RLHF).
We present two new post-training algorithms: (1) a rejection sampling fine-tuning algorithm with teacher committee (iTeC), and (2) a reinforcement learning from human feedback (RLHF) algorithm with mirror descent policy optimization and a leave-one-out advantage estimator (MDLOO) that are used on our reinforcement learning iterations and lead to significant model quality improvements.

\subsection{Data}

We use a hybrid data strategy in our post-training pipeline, which consists of both human annotated and synthetic data.
Throughout our data collection and experiment process, we have found data quality to be the key to model success and thus have conducted extensive data curation and filtering procedures.

\subsubsection{Human annotations}\label{sec:human_annotations}

\paragraph*{Demonstration data}
To fuel the instruction fine-tuning of \ourmodel{},
we collect high-quality human annotated demonstration datasets from various sources.
This dialogue-style data consists of both system-level and task-level instructions (a.k.a. prompts),
as well as their corresponding responses.
Similar to~\cite{zhou2024lima},
we observe quality to weigh more importantly than quantity in our experiments.
As a result,
we focus on key data quality criteria including helpfulness, harmlessness, presentation, and response accuracy,
in addition to targeting a diverse task distribution covering Apple Intelligence features.
To protect user privacy,
we take steps to verify no personally identifiable information is present in our data, and we do not include any personal data stored by users with Apple.

\paragraph*{Human preference feedback}
To iteratively improve \ourmodel{}'s capabilities,
we further collect human feedback for reinforcement learning.
In particular,
we instruct human annotators to compare and rank two model responses for the same prompt to collect side-by-side preference labels.
In addition, we also use single-side questions to guide this process.
These questions inform raters to grade the model response quality of various aspects including instruction following, safety, factuality, and presentation,
and we also retain these labels for model training.
We emphasize Apple values and standards in the process.
Similar to demonstration data,
we find data quality to be crucial for feedback data,
and thus we iterate data and model qualities jointly to improve them in a unified flywheel.

\subsubsection{Synthetic data}\label{sec:synthetic_data}

In addition to human annotations, we delve into enhancing data quality and diversity through synthetic data generation. Our findings suggest that when guided by our robust reward models, \ourmodel{}s are capable of generating high quality responses and for some specific domains, these responses are found to be on par with, or even superior to, human annotations. Therefore, we extend our prompt set to increase the diversity and find that those generated responses can benefit \ourmodel{}s themselves. In the following, we discuss three domains where we generate synthetic data for \ourmodel{} post-training: mathematics, tool use, and coding.

\paragraph*{Mathematics}
In the field of mathematics, the wide-ranging subjects and difficulty level make it  exceptionally resource-intensive for collecting human demonstrations, since it requires expert knowledge from the human writers. It also becomes impractical to solely rely on human-written content as the model continuously improves. As a consequence, exploring the potential of synthetic data becomes essential to effectively address the challenges. 

The creation of synthetic data for mathematics involves two primary stages: generating synthetic math problems and producing their corresponding solutions. For math problem synthesis, we employ several ``evolution" strategies where a seed set of prompts are transformed into a much larger set of diverse prompts: 

    \begin{itemize}
        \item[] \textbf{Problem rephrase and reversion.} Following the approach in~\cite{yu2023metamath}, we prompt \ourmodel{} to rephrase seed math questions, and curate reverse questions to derive a specific number in a raw problem statement when provided with the final answer.    
        \item[] \textbf{Problem evolution.} Inspired by the instruction evolving technique~\cite{xu2023wizardlm}, given a seed problem set $\mathcal{D}_\text{seed}$ we prompt \ourmodel{} to generate two distinct sets of math problems, i.e.
        $F(\mathcal{D}_{\text{seed}}) \xrightarrow[\text{}]{\scriptstyle \text{depth}} \mathcal{D}_{\text{depth}}$, and $F(\mathcal{D}_{\text{seed}}) \xrightarrow[\text{}]{\scriptstyle \text{breadth}} \mathcal{D}_{\text{breadth}}$. The in-depth evolution enhances instructions by adding complexities while the in-breadth evolution improves the topic coverage.
        For both $\mathcal{D}_\text{breadth}$ and $\mathcal{D}_\text{depth}$, we first perform de-duplication with an embedding model, and subsequently prompt LLMs to ensure the coherence and solvability of the math problems. In addition, for $\mathcal{D}_\text{depth}$ a difficulty level is assigned and we only select math problems that score above a specified threshold.
    \end{itemize}

With an augmented set of math questions, we then prompt \ourmodel{} to synthesize $N$ responses with chain-of-thought per question. If the initial seed data has ground truth, they can be used as an ``outcome reward signal'' to filter synthesized answers.
For problems that require less reasoning steps, we observe that a correct final answer often gets associated with correct intermediate steps. If direct answer checking is unsuccessful or ground truth is unavailable, we instead assess the response correctness by querying an LLM judge. We find that the filtered answers, when fed into the training data, boost our models' math capabilities by a large margin.

\paragraph*{Tool use}
We develop tool-use capabilities such as function call, code interpreter, and browsing through a mixture of synthetic and human data.
The model capabilities are first bootstrapped with synthetic data, which focuses on single-tool use cases.
We then collect human annotations to improve model capabilities that involve multi-tool and multi-step scenarios.
We further augment the human curated function call data by mixing the oracle tool with other similar tools to increase the difficulty of tool selection.
In addition, we synthesize parallel function call from human curated function call data to enable the new capability and tool intent detection data based on human curated function call and general SFT data to mitigate tool call over-triggering issues. 

\paragraph*{Coding}
The generation of a synthetic coding dataset involves a self-instruct method with rejection sampling. This approach enables the model to learn and generate data autonomously. Starting with 71 different programming topics as the seeds, the model is prompted to generate an initial pool of coding interview-like questions. For each question, the model generates a set of unit tests and a number of potential solutions. We then use an execution-based rejection sampling method to select the best solution. This involves compiling each potential solution with every unit test and executing them. The solution with the highest number of successful executions is chosen. This results in a collection of (question, test cases, solution) triplets. At the end, we validate the quality of the dataset by filtering the triplets using the number of passed unit tests, resulting in 12K high quality triplets used in the SFT.

\subsection{Supervised fine-tuning (SFT)}

It has been shown~\cite{chung2024scaling} that scaling multi-task instruction tuning dramatically enhances model performance on a wide variety of tasks. Similarly, we attempt to scale supervised fine-tuning data to achieve a strong base model for subsequent alignment. During SFT, we collect and train models on demonstration data of a given prompt\footnote{A prompt may consist of the most recent user instruction as well as all previous user-model-system interactions.}.
We carefully select and combine both human data and synthetic data to form a high quality mixture that covers various natural language use cases.

\paragraph{Data selection} We establish a series of quality guards of the data before onboarding them for model training, including ratings from in-house human labelers, automatic model-based filtering techniques, and deduplication with text embeddings. We also scale up the mixture size by a variety of synthetic data generation methods, as described in \autoref{sec:synthetic_data}, and rejection sampling as described in \autoref{sec:teaching_committee}.

\paragraph{Tuning the mixture ratio} In order to tune the mixture weight, we treat it as an optimization problem. Specifically, given a set of weights $(w_1, w_2, ..., w_n)$ where $w_i$ represents the ratio of a specific component in the mixture, we train a model with $w_i \rightarrow w_i \pm \Delta w_i$ and evaluate the quality change on a set of benchmarks.
We find that extensively running such experiments can effectively identify the best mixture and remove the least impactful data components.

\paragraph{Training hyperparameters} The model is trained with a constant learning rate $5\mathrm{e}{-6}$ for \ourmodel{-server} and $2\mathrm{e}{-5}$ for \ourmodel{-device} models, as well as a drop out rate $0.1$. Since the evaluation metrics fluctuate across different checkpoints, we run checkpoint selection based on automatic evaluation benchmarks and best-of-N selection with reward models to test the headroom for RL. 

\subsection{Reinforcement learning from human feedback (RLHF)}

We further use reinforcement learning with collected human preference data to improve model performance and quality.
This involves training a robust reward model and applying it in two algorithms of iTeC and MDLOO that we discuss below. We describe more details of our RLHF pipeline in \autoref{sec:appendix_hlhf}.

\subsubsection{Reward modeling}

We train reward models using the human preference data collected with the method in \autoref{sec:human_annotations}. 
Each human preference data item contains one prompt and two responses along with human labels including:

\begin{itemize}[itemsep=0pt]
    \item The preferred response between the two and the preference level, i.e., whether the preferred response is significantly better, better, slightly better, or negligibly better than the rejected response.
    \item The single-sided grading of each response, measuring the instruction following property, the conciseness, truthfulness, and harmlessness of each of the responses.
\end{itemize}

Our reward model training follows the standard practice of reward modeling in RLHF with two main innovations:

\begin{itemize}[itemsep=0pt]
    \item We design a soft label loss function that takes the level of human preference into account.
    \item We incorporate single-sided gradings as regularization terms in reward modeling.
\end{itemize}

We employ the commonly used Bradley-Terry-Luce (BTL) model~\cite{bradley1952rank} for reward modeling in RLHF. In this model, the probability that a human annotator prefers one response over another is modeled as the sigmoid function of the difference of the rewards. Our \emph{soft label} loss function encourages that this probability is high when the preference level is high, e.g., when one response is significantly better than the other, and vice versa. We note that this is different from the \emph{margin}-based loss function in Llama 2~\cite{llama2}, which also leverages the preference level. Empirically, we find that our method works better than the margin-based loss function. Moreover, we also find that using the single-sided gradings as regularization terms can effectively improve the accuracy of the reward model. More details of our reward modeling techniques can be found in \autoref{sec:appendix_reward_model}.

\subsubsection{\committeefull{} (\committee{})}\label{sec:teaching_committee}

To fully unlock the ability of our model with multiple rounds of RLHF,
we propose a novel iterative RLHF framework which effectively combines various preference optimization algorithms, 
including rejection sampling (RS), Direct Preference Optimization (DPO)~\cite{dpo} and its variants such as IPO~\cite{ipo}, and online reinforcement learning (RL). 
This enables us to bring the benefit of RLHF to \ourmodel{} models across all sizes and improve their alignment at the same time.

\paragraph{Iterative committee}
One of the most important lessons we learned from developing \ourmodel{} RLHF
is to refresh online human preference data collection using a diverse set of the best performing models.
Specifically, for each batch of human preference data collection,
we set up a collection of latest promising models trained from SFT, RS, DPO/IPO, and RL,
as well as best models from the previous iterations,
which we refer to as ``model committee''.
We collect pairwise human preference on responses sampled from the latest model committee.

After acquiring each batch of human preference data,
we refresh our reward model, and further train a new set of models
using the collection of preference optimization algorithms.
We then continue the next round of iterative RLHF data collection with a new model committee.

\paragraph{Committee distillation}
We further run rejection sampling (distillation) from the model committee with the latest reward model as a reranker.
Instead of reranking at global-level, i.e., picking a single best performing model from the committee and using it as a teacher model,
we rerank model responses at the prompt-level.
Specifically, for each prompt,
we sample multiple responses from each model in the committee,
and use the latest reward model to select the best response for each prompt.
This allows us to combine the advantages of models trained by different preference optimization algorithms. For instance,
we find that algorithms that leverage negative examples, e.g., online RLHF, DPO, IPO, to be better in improving reasoning skills such as math, while rejection sampling fine-tuning learns instruction following and writing skills more effectively.

\paragraph{Scaling up distillation}
In order to bring the RLHF improvements to \ourmodel{} models across all sizes,
we scale up distillation from the model committee.
Different from larger models, where carefully iterating data and model quality matters much more than data quantity,
we find smaller models can achieve tremendous improvement when we scale up the number of prompts for distillation.
Our final \ourmodel{-on-device} model is trained on more than 1M high quality responses generated from the model committee.

\subsubsection{Online RLHF algorithm: MDLOO}
\label{sec:rl}

In this section, we introduce our online reinforcement learning algorithm MDLOO,
where we decode responses during model training and apply RL algorithms to maximize the reward.

We use the commonly adopted RLHF objective that maximizes the KL-penalized reward function~\cite{ouyang2022training}:
\begin{equation}\label{eq:rl_objective}
    \max_{\theta} \mathbb{E}_{x \sim \mathcal{D}, y \sim \pi_\theta(\cdot|x)} \left[ r_\phi(x, y) - \beta D_{\text{KL}} \left( \pi_\theta(\cdot|x) \| \pi_{\text{ref}}(\cdot|x) \right) \right],
\end{equation}
where $\mathcal{D}$ is the prompt distribution, $D_{\text{KL}}(\cdot \| \cdot)$ denotes the Kullback-Leibler divergence between two distributions, and $\beta$ is the coefficient that controls the divergence between the behavior policy $\pi_{\theta}$ and a reference policy $\pi_{\text{ref}}$, which is typically a model trained by SFT. In our RL training, we use the reward function
\begin{align}\label{eq:reward}
R(x, y) = r_\phi(x, y) - \beta \log\frac{ \pi_{\theta}(y|x) }{ \pi_{\text{ref}}(y|x) },
\end{align}
whose expectation is equivalent to \autoref{eq:rl_objective}. We consider the bandit setting where the generation of the entire response is considered as one action, and we do not use the value network (a.k.a. the critic) to obtain the per-token reward or advantage.

Similar to commonly used RLHF algorithms such as PPO~\cite{ppo:schulman2017proximal}, we use a trust-region based policy iteration algorithm.
We made two main design choices in our online RL algorithm:

\begin{itemize}[itemsep=0pt]
\item We use the Leave-One-Out (LOO) estimator to estimate the advantage of a prompt-response pair, similar to a recent work~\cite{ahmadian2024back}.
\item We use Mirror Descent Policy Optimization (MDPO)~\cite{mdpo:tomar2020mirror} to optimize the policy, differently from the more commonly used clipping-based PPO method.
\end{itemize}

Thus, we name our online RL algorithm \emph{Mirror Descent with Leave-One-Out estimation} (MDLOO). More specifically, during the decoding stage of the algorithm, we decode multiple responses for each prompt, and assign the \emph{advantage} of each response to be the difference of the reward of the (prompt, response) pair and the mean reward of the other responses generated by the same prompt. Intuitively, this estimator aims to measure how much better a response is compared to a typical response. Empirically, we find this advantage estimator crucial for stabilizing the RL algorithm and achieving strong results.
Moreover, we use a KL-regularization-based trust region method, i.e. MDPO, to control the policy change in each iteration. We find that this algorithm is more effective than PPO in our setting. More details of our online RLHF algorithm can be found in \autoref{sec:appendix_rl}.

\section{Powering Apple Intelligence features}
\label{sec:features}

Our foundation models are designed for Apple Intelligence, the personal intelligence system integrated into supported models of iPhone, iPad, and Mac.
We have built these models to be fast and efficient. And while we have achieved impressive levels of broad capability in our base model, the actual relevant measure of its quality is how it performs on specific tasks across our operating systems.

\begin{figure}[h]
    \centering
    \includegraphics[width=1.2\textwidth, alt={Architecture Diagram of Apple Intelligence.},center]{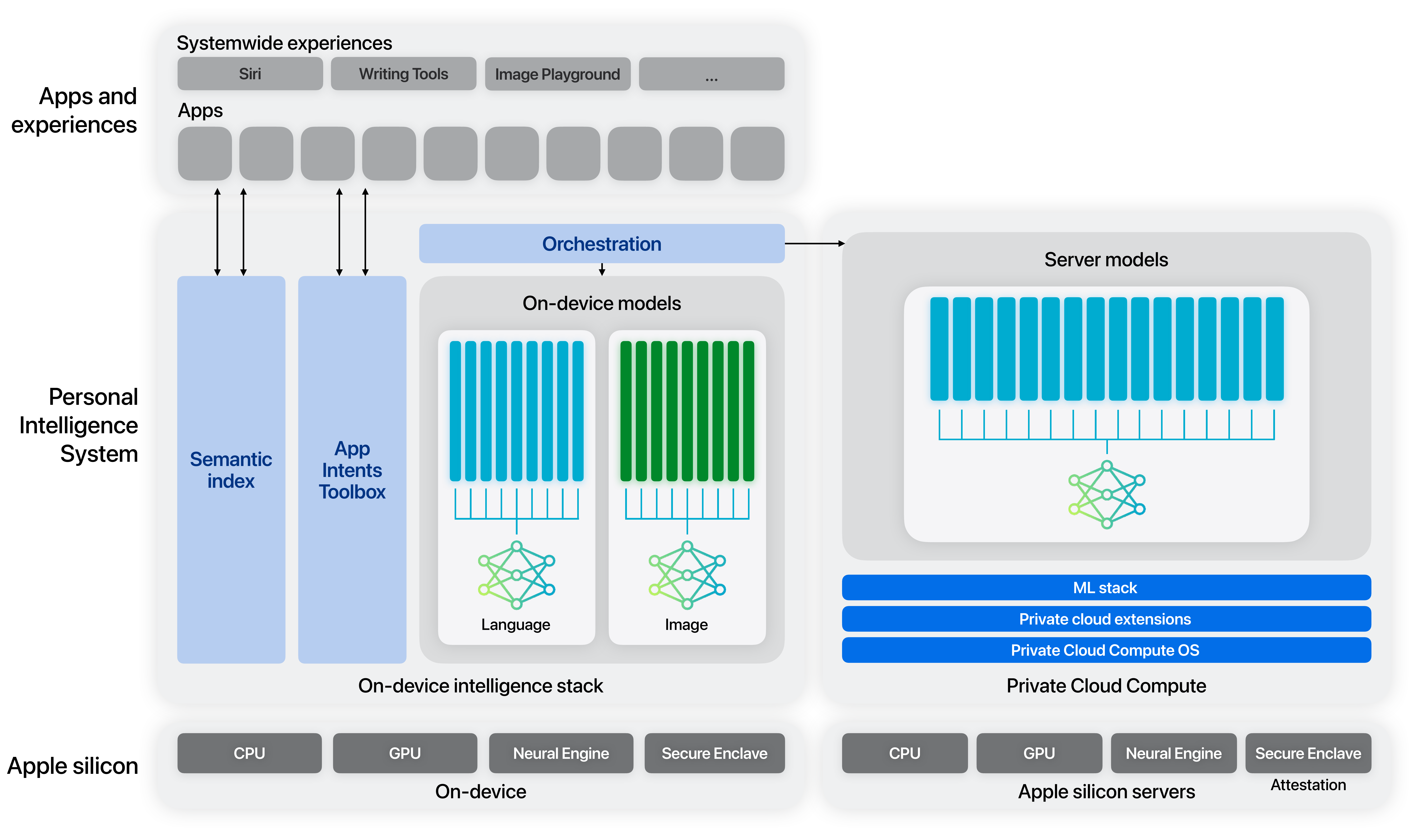}
    \caption{Architecture of Apple Intelligence with adapters for the language on-device and server models and the image models. In this report we are only describing the text models.}
    \label{fig:architecture}
\end{figure}

Here we have found that we can elevate the performance of even small models to best-in-class performance through task-specific fine-tuning and have developed an architecture, based on runtime-swappable adapters, to enable the single foundation model to be specialized for dozens of such tasks. A high-level overview is presented in \autoref{fig:architecture}.

\subsection{Adapter architecture}
Our foundation models are fine-tuned for users' everyday activities, and can dynamically specialize themselves on-the-fly for the task at hand. We use LoRA~\cite{hulora} adapters, small neural network modules that can be plugged into various layers of the base model, to fine-tune our models for specific tasks.
For each task, we adapt all of \ourmodel{}'s linear projection matrices in the self-attention layers and the fully connected layers in the pointwise feedforward networks.
By fine-tuning only the adapters, the original parameters of the base pre-trained model remain unchanged, preserving the general knowledge of the model while tailoring the adapters to support specific tasks.

We represent the values of the adapter parameters using 16 bits, and for the $\sim$3 billion parameter on-device model, the parameters for a rank 16 adapter typically require 10s of megabytes. The adapter models can be dynamically loaded, temporarily cached in memory, and swapped---giving our foundation model the ability to specialize itself on the fly for the task at hand while efficiently managing memory and guaranteeing the operating system's responsiveness.

To facilitate the training of the adapters, we created an efficient infrastructure that allows us to rapidly add, retrain, test, and deploy adapters when the base model or the training data gets updated or new capabilities are required. It is worth noting that the adapter parameters are initialized using the accuracy-recovery adapter introduced in \autoref{sec:optimizations}.

\subsection{Optimizations} %
\label{sec:optimizations}
The \ourmodel{} models are designed to support our users throughout their daily activities, and both inference latency and power efficiency are important for the overall user experience.
We apply various optimization techniques to allow \ourmodel{} to be efficiently deployed on-device and in Private Cloud Compute. These techniques significantly reduce memory, latency, and power usage while maintaining the overall model quality.

\paragraph{} In order to fit \ourmodel{} into a constrained memory budget of edge devices and reduce inference cost, it is critical to apply model quantization techniques to reduce the effective bits per weight while maintaining the model quality. Previous works have found that 4-bit quantized models only have marginal loss of quality (typically measured in pre-training metrics) compared to the original 32/16-bit float-point versions. Since \ourmodel{} is expected to support a diverse set of product features, it is essential that the quantized model retains capabilities in specific domains critical to these use cases. To achieve an optimal trade-off between model capacity and inference performance, we have developed state-of-the-art quantization methods and a framework that utilizes accuracy-recovery adapters. This allows us to achieve near-lossless quantization that is on average less than 4 bit-per-weight, and provides flexible quantization scheme choices.

\paragraph{Methods} The model is compressed and quantized, on average under 4-bit-per-weight, after the post-training stages (details of the quantization scheme will be discussed later). The quantized model often shows a moderate level of quality loss. Therefore, instead of directly passing the quantized model to application teams for feature development, we attach a set of parameter-efficient LoRA adapters for quality recovery.  We make sure that these LoRA adapters training recipes are consistent with  pre-training and post-training processes. Then, products will fine-tune their own feature-specific LoRA adapters by initializing the adapter weights from the accuracy-recovery adapters, while keeping the  quantized base model frozen.

It is noteworthy that training accuracy-recovery adapters is sample-efficient and can be considered as a mini-version of training the base model. During the pre-training stage of the adapters, we only require approximately 10 billion tokens ($\sim0.15\%$ of base model training) to fully recover the capacity for the quantized model. Since application adapters will fine tune from these accuracy-recovery adapters, they do not incur any additional memory usage or inference costs. Regarding adapter size, we found that adapter rank of 16 offers the optimal tradeoff between model capacity and inference performance. However, to provide flexibility for various use cases, we provide a suite of accuracy-recovery adapters in different ranks $\{8,16,32\}$ for application teams to select from. In \autoref{sec:quant_ablation}, we provide detailed evaluation results across unquantized, quantized, and accuracy-recovered models and show that the recovered models perform much closer to the unquantized version.

\paragraph{Quantization schemes}
Another benefit brought by accuracy-recovery adapters is that they allow more flexible choices of quantization schemes. Previously when quantizing LLMs, people typically group the weights into small blocks, normalize each block by the corresponding maximal absolute values to filter out outliers, then apply quantization algorithms in a block basis. While a larger block size yields lower effective bits per weight and a higher throughput, the quantization loss would increase. In order to balance this tradeoff, it is common to set block size to a small value, like 64 or 32. In our experiments, we found that accuracy-recovery adapters can greatly improve the pareto frontier in the tradeoff. More errors will be recovered for more aggressive quantization schemes. As a result, we are able to use a highly-efficient quantization scheme for \ourmodel{} without worrying about losing model capacity. Specifically, our \ourmodel{-on-device} model running on Apple Neural Engine (ANE) uses palettization: for projection weights, every 16 columns/rows share the same quantization constants (i.e., lookup tables) and are quantized using K-means with 16 unique values (4-bit). The quantization block size can be up to 100k. Besides, since \ourmodel{'s} embedding layer is shared between the input and output, it is implemented differently from projection layers on ANE. Hence, we quantize the embedding using per-channel quantization with 8-bit integers for better efficiency.

\paragraph{Mixed-precision quantization} Residual connections exist in every transformer block and every layer in \ourmodel{}. So it is unlikely that all layers have the equal importance. Following this intuition, we further reduce the memory usage by pushing some layers to use 2-bit quantization (default is 4-bit). On average, \ourmodel{-on-device} can be compressed to only about 3.5 bits per weight (bpw) without significant quality loss. We choose to use 3.7 bpw in production as it already meets the memory requirements.

\paragraph{Interactive model analysis}
We use an interactive model latency and power analysis tool, Talaria~\cite{hohman2024talaria}, to better guide the bit rate selection for each operation.

\paragraph{More discussions} The usage of quantized model and LoRA adapters look conceptually similar to QLoRA~\cite{dettmers2024qlora}.
While QLoRA was designed to save computational resources during fine-tuning, our focus is on the ability to switch between different LoRA adapters to efficiently support high performance across various specific use cases.
Before feature-specific finetuning, we first train accuracy-recovery adapters on the same pretraining and post-training data, which is critical to preserve the model quality.
The accuracy-recovery framework can be combined with different quantization techniques, like GPTQ~\cite{frantar2022gptq} and AWQ~\cite{lin2024awq}, since it does not depend directly on the quantization method itself.
The feature adapters described in \autoref{sec:features} are initialized from these accuracy-recovery adapters.

\subsection{Case study: summarization}  %
We use the \ourmodel{-on-device} model to power summarization features. We worked with our design teams to create specifications for summaries of Emails, Messages, and Notifications. 

\paragraph{} While \ourmodel{-on-device} is good at general summarization, we find it difficult to elicit summaries that strictly conform to the specification. Therefore, we fine tune a LoRA adapter on top of the quantized \ourmodel{-on-device} for summarization. The adapter is initialized from the accuracy-recovery adapter as described in \autoref{sec:optimizations}. 
We use a data mixture consisting of input payloads covering Emails, Messages, and Notifications. These payloads include public dataset, vendor data, and internally generated and submitted examples. All the data have been approved to use for production. Vendor data and internally generated data have been anonymized to remove the user information.
Given these payloads, we generated synthetic summaries using \ourmodel{-server} according to product's requirements. These payloads and summaries are used for training. 

\paragraph{Synthetic summaries}
We use \ourmodel{-server} to generate synthetic summaries. We apply a series of rule-based filters followed by model based filters. Rule-based filters are based on heuristics such as length constraints, formatting constraints, points of view, voice, etc. Model-based filters are used to screen more challenging problems such as entailment.
Our synthetic data pipeline allows us to efficiently generate a large amount of training data and filter it out by an order of magnitude to retain high-quality examples for fine tuning.

\paragraph{Prompt injection}
We find that \ourmodel{-on-device} is prone to following instructions or answering questions that are present in the input content instead of summarizing it. To mitigate this issue, we identify a large set of examples with such content using heuristics, use \ourmodel{-server} to generate summaries, as it does not exhibit similar behavior, and add this synthetic dataset to the fine tuning data mixture.

\section{Evaluation}

We evaluate the \ourmodel{} models on pre-training (\autoref{sec:eval_pretrain}), post-training (\autoref{sec:eval_posttrain}), and most importantly, feature-specific (\autoref{sec:eval_feature}) benchmarks.

\subsection{Pre-training evaluation}
\label{sec:eval_pretrain}
In this section we present common few-shot pre-training evaluation metrics.
While these benchmarks are useful for tracking our progress on pre-training, we found that human evaluations on the post-trained models (\autoref{sec:eval_posttrain}) and feature adapters (\autoref{sec:eval_feature}) are more closely correlated to end-to-end user experience.

\paragraph{} We evaluate \ourmodel{} pre-trained models with common open-sourced evaluation harnesses and benchmarks.
\autoref{table:helm-mmlu} presents the results of \ourmodel{-on-device} and \ourmodel{-server} on HELM MMLU v1.5.0~\cite{helm2023}, which tests 5-shot multiple-choice question answering across 57 subjects.
Also see \autoref{table:openllmleaderboard} and \autoref{table:helmlite} for the results of \ourmodel{-server} on a subset of the HuggingFace OpenLLM leaderboard V1~\cite{openllmharness} and the HELM-Lite v1.5.0 benchmark suite~\cite{helm-lite-1.5.0}, respectively. 
These benchmarks show that the \ourmodel{} pretrained models possess strong language and reasoning capabilities and provide a solid foundation for post-training and feature fine-tuning.

\begin{table}[!h]
    \centering
    \small
    \begin{tabular}{@{}l>{\raggedleft\arraybackslash}cc@{}}
        \toprule
        \textbf{} & \textbf{\ourmodel{-on-device}} & \textbf{\ourmodel{-server}} \\ \midrule
        \textbf{MMLU (5 shot)} & 61.4 & 75.4 \\ 
        \bottomrule
    \end{tabular}
    \caption{HELM MMLU-5s~\cite{helm2023} v1.5.0 evaluation results.}
    \label{table:helm-mmlu}
\end{table}

\begin{table}[!h]
    \centering
    \small
    \begin{tabular}{@{}l>{\raggedleft\arraybackslash}r@{}}
        \toprule
        \textbf{} & \textbf{AFM-server} \\ \midrule
        \textbf{MMLU (5-shot)} & 75.3 \\ 
        \textbf{GSM8K (5-shot)} & 72.4 \\ 
        \textbf{ARC-c (25-shot)} & 69.7 \\ 
        \textbf{HellaSwag (10-shot)} & 86.9 \\ 
        \textbf{Winogrande (5-shot)} & 79.2 \\ 
        \bottomrule
    \end{tabular}
    \caption{A subset of Open LLM Leaderboard~\cite{openllmharness} V1 evaluation results.}
    \label{table:openllmleaderboard}
\end{table}

\begin{table}[!h]
    \centering
    \small
    \begin{tabular}{@{}l>{\raggedleft\arraybackslash}r@{}}
        \toprule
        & \textbf{AFM-server} \\ \midrule
        \textbf{Narrative QA} & 77.5 \\ 
        \textbf{Natural Questions (open)} & 73.8 \\ 
        \textbf{Natural Questions (closed)} & 43.1 \\ 
        \textbf{Openbook QA} & 89.6 \\ 
        \textbf{MMLU} & 67.2 \\ 
        \textbf{MATH-CoT} & 55.4 \\
        \textbf{GSM8K} & 72.3  \\ 
        \textbf{LegalBench} & 67.9 \\ 
        \textbf{MedQA} & 64.4  \\ 
        \textbf{WMT 2014} & 18.6  \\ \bottomrule
    \end{tabular}
    \caption{HELM-Lite~v1.5.0~\cite{helm-lite-1.5.0} pre-training evaluation results. N.B. Many benchmarks (e.g. MMLU) differ significantly from commonly used settings.}
    \label{table:helmlite}
\end{table}

\subsection{Post-training evaluation}
\label{sec:eval_posttrain}

We evaluate post-training models on comprehensive benchmarks and 
compare \ourmodel{} models with various open-source models, as well as GPT-3.5 and GPT-4.\footnote[1]{
We compared against the following model versions: gpt-3.5-turbo-0125, gpt-4-0125-preview, Gemini‑1.5‑Pro‑0514, DBRX Instruct, Phi-3-mini-4k-instruct, LLaMA 3 8B Instruct, LLaMA 3 70B Instruct, Mistral-7B-Instruct-v0.2, Mixtral-8x22B-Instruct-v0.1, Gemma-1.1-2B, and Gemma-1.1-7B.}
All results reported in this section are obtained using \ourmodel{-on-device} and \ourmodel{-server} base models without any adapter, in \texttt{bfloat16} precision. In this section, we first present human evaluation results that measure the \ourmodel{}s general capabilities, and then present results for several specific capabilities and domains.

\subsubsection{Human evaluation}\label{sec:human_eval}

Human evaluation simulates practical use cases and user feedback, and so often serves as the gold standard for language model evaluation.
Consequently, we conduct extensive human evaluations both while developing the model and to evaluate its final form.
We collect sets of evaluation prompts to test the model on different aspects, including both general capabilities and safety.
For each prompt, two model responses are presented to human raters anonymously for side-by-side comparisons.
Depending on the nature of the evaluation,
a detailed guideline containing grading principles and examples of single-response ratings and side-by-side preference ratings is provided to human raters to ensure consistent grading standards and evaluation quality.
Each pair of model responses is graded by multiple graders and their ratings are aggregated for final results.
Overall, we find human evaluation to align better with user experience and provide a better evaluation signal than some academic benchmarks that use LLMs as graders. In this section, we present the results for human evaluation on general capabilities, and the safety evaluation results are provided in \autoref{sec:safety-eval}.

We collect a comprehensive set of $1393$ prompts to evaluate the general model capabilities.
These prompts are diverse across different difficulty levels and cover major categories including:
analytical reasoning,
brainstorming,
chatbot,
classification,
closed question answering,
coding,
extraction,
mathematical reasoning,
open question answering,
rewriting,
safety,
summarization,
and writing. 
To prevent overfitting, when preparing training data, we conduct decontamination against our evaluation prompts.

\inputhide{tables/human_eval_main}
\begin{figure}[!ht]
    \centering
     \includegraphics[width=\textwidth, alt={Chart showing which models win, tie, or loose in a direct comparison. On-device comparisons are at the top and server comparisons are at the bottom.},center]{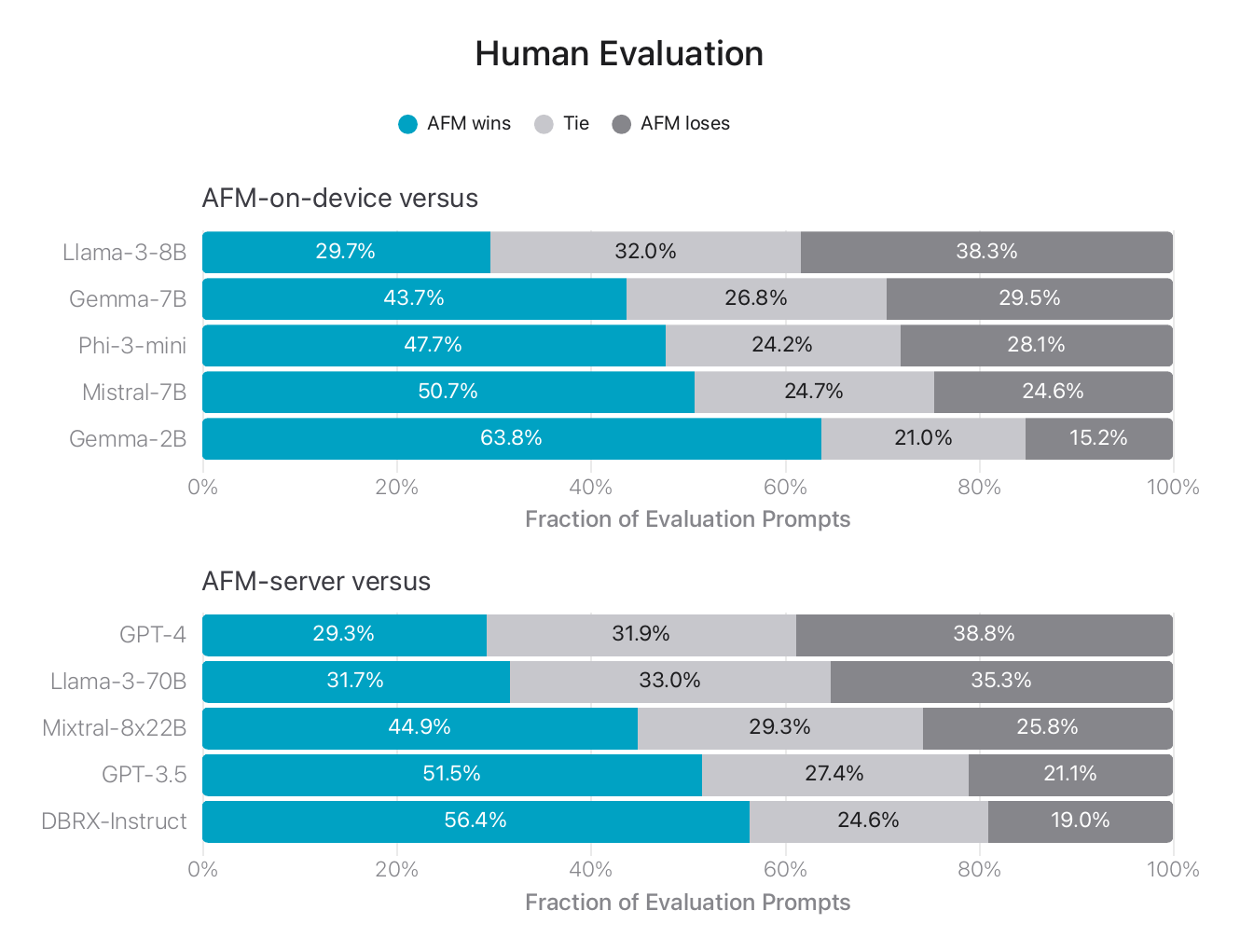}
    \caption{Side-by-side evaluation of \ourmodel{-on-device} and \ourmodel{-server} against comparable models. We find that our models are often preferred over competitor models by human graders.}
    \label{fig:human-eval}
\end{figure}

In \autoref{fig:human-eval}, %
we compare \ourmodel{} with both open-source models (Phi-3, Gemma-1.1, Llama-3, Mistral, DBRX-Instruct) and commercial models (GPT-3.5, and GPT-4).
\ourmodel{} models are preferred by human graders over competitor models.
In particular, \ourmodel{-on-device} obtains a win rate of 47.7\% when compared to Phi-3-mini despite being 25\% smaller in model sizes,
and even outperforms open-source strong baselines Gemma-7B and Mistral-7B that are more than twice larger in the number of parameters.
When compared to closed-source models, \ourmodel{-server} achieves competitive performance, scoring a win rate of more than $50\%$ and a tie rate of 27.4\% against GPT-3.5.

\subsubsection{Instruction following}

Instruction following (IF) is the core capability we desire of language models,
as real-world prompts are often sophisticated and contain complex instructions.
We emphasize the importance of instruction following in both our RLHF data collection and human evaluation.
In this subsection, we evaluate our models' IF skills using automated benchmarks.

\begin{figure}[!ht]
    \centering
     \includegraphics[width=1.1\textwidth, alt={Chart with bars for the porformance of various models on various benchmarks. On-device models are on the left and server models are on the right.},center]{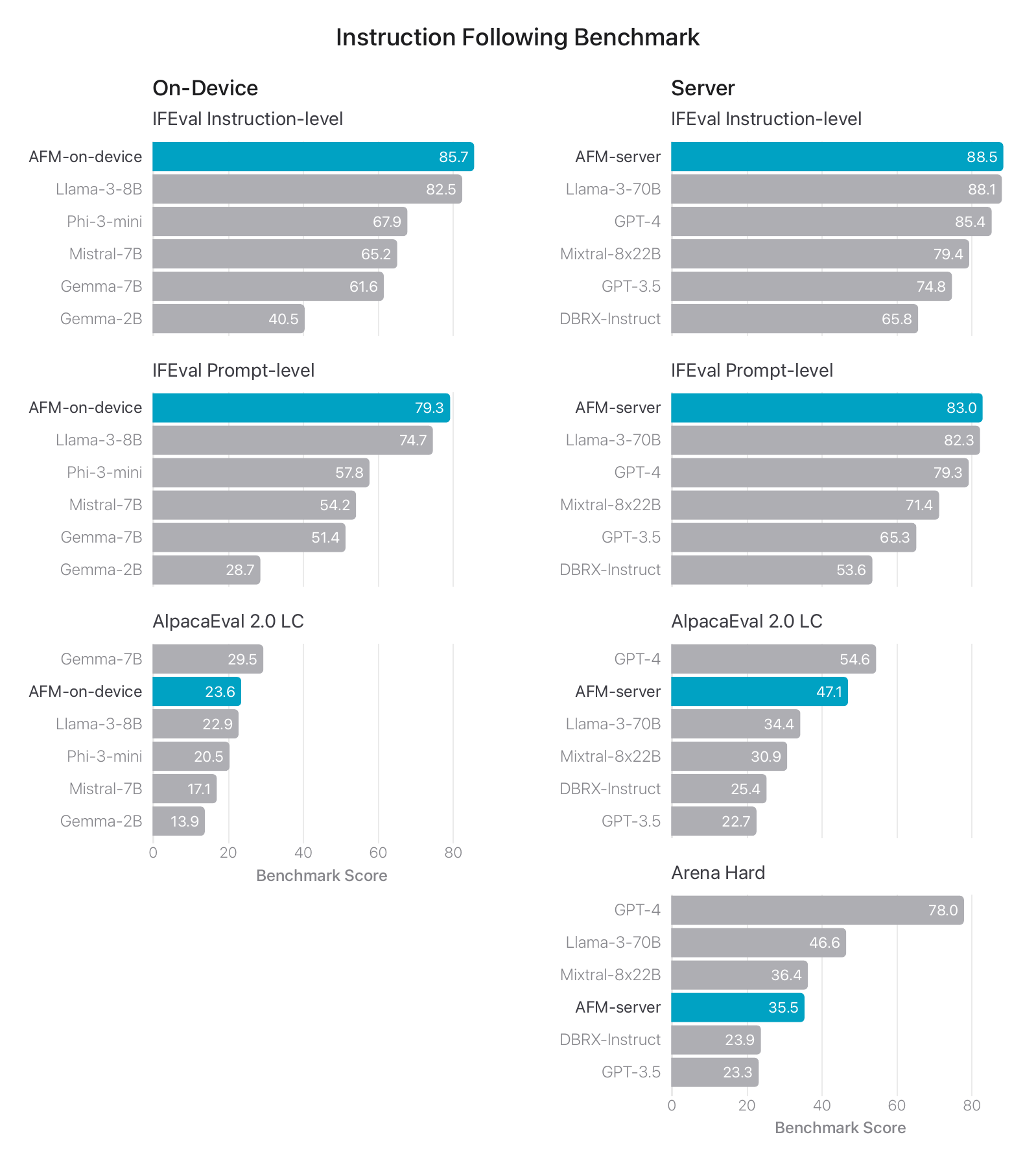}
    \caption{Instruction-following capability (measured with IFEval) for \ourmodel{} models and relevant comparison models (higher is better). The AlpacaEval~2.0~LC results for Mistral~7B, Llama3~8B, Llama3~70B, DBRX-Instruct, and Mixtral~8x22B are obtained from \href{https://tatsu-lab.github.io/alpaca_eval/}{the AlpacaEval leaderboard}~\cite{alpaca}. The Arena Hard results for comparison models are from \href{https://github.com/lm-sys/arena-hard-auto}{the Arena-Hard-Auto leaderboard}~\cite{arenahard2024}. All other results are from our own evaluations.}
    \label{fig:instruction-following}
\end{figure}

In \autoref{fig:instruction-following}
we evaluate \ourmodel{-on-device} and \ourmodel{-server} on the public IFEval benchmark~\cite{zhou2023instruction}, respectively.
This benchmark measures a language model's capability to generate responses that precisely follow  instructions in the prompt. The instructions in this benchmark typically include requirements on the response length, format, content, etc.
We find \ourmodel{-on-device} and \ourmodel{-server} to achieve superior performance on both instruction-level and prompt-level accuracy.
In addition,
we also benchmark \ourmodel{} models on the AlpacaEval 2.0 LC benchmark~\cite{dubois2024length} to measure general instruction-following capability, and results suggest that our models are highly competitive.

\inputhide{tables/if_eval}

\subsubsection{Tool use}
In tool use applications, 
given a user request and a list of potential tools with descriptions, 
the model can choose to issue tool calls by providing a structured output specifying the name and parameter values of the tools to call. We expect the tool descriptions to follow the OpenAPI specification.\footnote{https://github.com/OAI/OpenAPI-Specification}

We evaluate on the public Berkeley Function Calling Leaderboard benchmarks~\cite{patil2023gorilla}
via native support of function calling, using the AST metrics.

\begin{figure}[!ht]
    \centering
     \includegraphics[width=1.1\textwidth, alt={Chart with bars comparing the score of various models on five function calling benchmarks.},center]{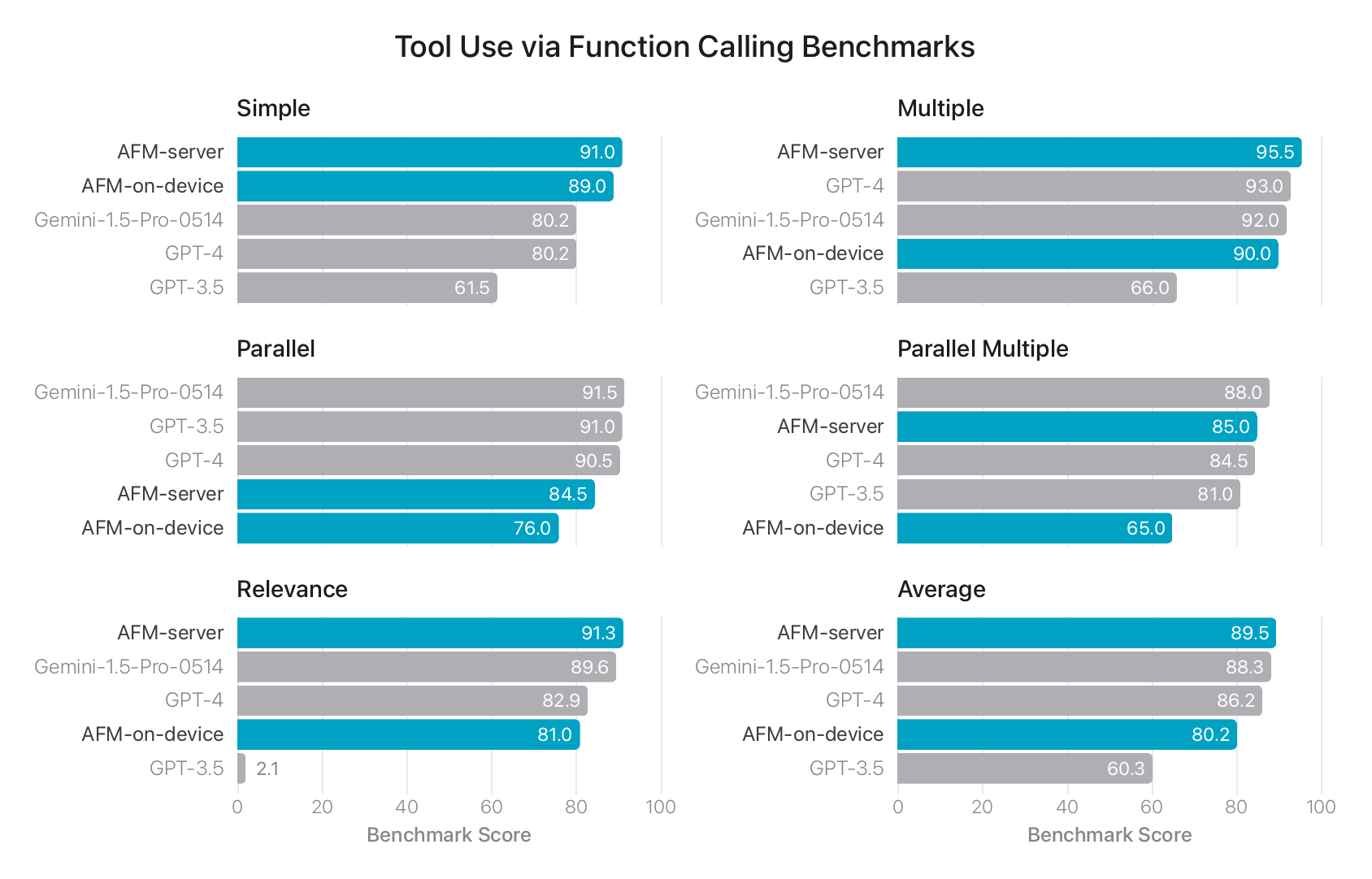}
    \caption{
    Berkeley Function Calling Leaderboard Benchmark
    evaluation results on Function Calling API, along-side relevant sampled comparisons.
    Numbers were collected from \href{https://huggingface.co/spaces/gorilla-llm/berkeley-function-calling-leaderboard/blob/e8e0a803d621a067f515ea4a0b970240a9fc19ed/data.csv}{the Gorilla leaderboard}~\cite{patil2023gorilla}.
    }
    \label{fig:tool-use}
\end{figure}

As shown in \autoref{fig:tool-use}, \ourmodel{-server} achieves the best overall accuracy, outperforming Gemini-1.5-Pro-Preview-0514 and GPT-4.

\inputhide{tables/tool_use_eval}

\subsubsection{Writing}

Writing is one of the most critical abilities for large language models to have, as it empowers various downstream use cases such as changing-of-tone, rewriting, and summarization.
However, assessing writing quality is a non-trivial task,  and not well-covered in the above public benchmarks.

We evaluate \ourmodel{}'s writing ability on our internal summarization and composition benchmarks, 
consisting of a variety of writing instructions.
Following LLM-as-a-judge~\cite{zheng2024judging}, we design a grading instruction
for each summarization and composition task,
and prompt GPT-4 Turbo to assign a score from 1 to 10 for model responses.\footnote{
Due to the choice of using GPT-4 as judge, the score of GPT-4 Turbo can be overestimated.
}
We note that there are certain limitations and biases associated with using an LLM as a grader, such as length bias.

\begin{figure}[!ht]
    \centering
     \includegraphics[width=1.1\textwidth, alt={Chart with bars showing the performance of various models for summarization and generation benchmarks. On-device models are on the left and server models are on the right.},center]{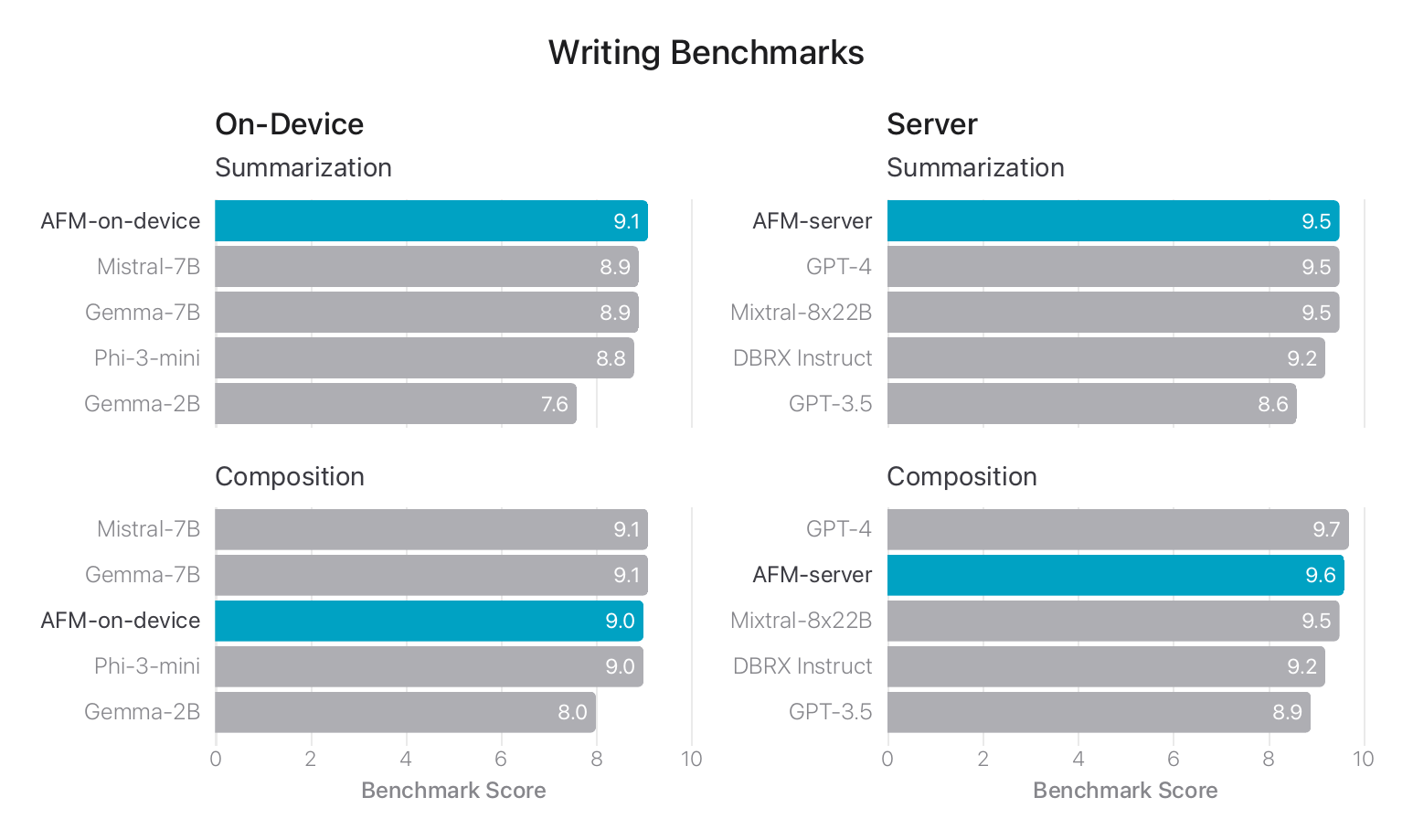}
    \caption{Writing ability on internal summarization and composition benchmarks (higher is better) for \ourmodel{-on-device} and \ourmodel{-server} alongside
relevant sampled comparisons. We find that our models perform better or similar to related models.}
    \label{fig:writing}
\end{figure}

We compare \ourmodel{} with a few of the most outstanding models,
along with smaller-scale open-source models.
As shown in \autoref{fig:writing},
\ourmodel{-on-device} can achieve comparable or superior performance when compared to \gemmal{} and \mistral{}.
\ourmodel{-server} significantly outperforms \dbrx{}{} and \gpts{} and is comparable to \gptl{}.

\inputhide{tables/writing_eval.tex}

\subsubsection{Math}

\begin{figure}[!ht]
    \centering
     \includegraphics[width=1.1\textwidth, alt={Chart with bars showing the performance of various models for GSM8K and MATH benchmarks. On-device models are on the left and server models are on the right.},center]{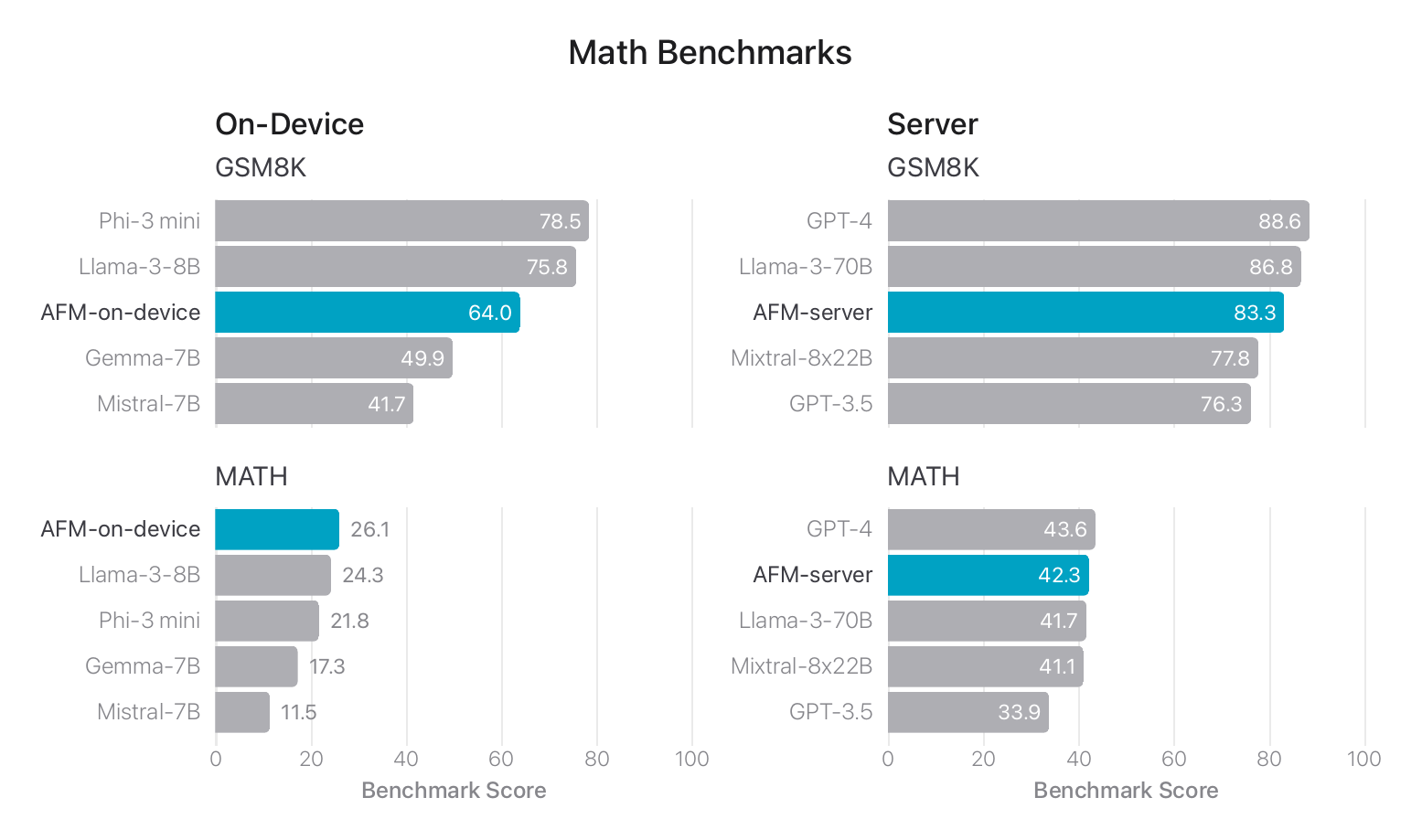}
    \caption{Math benchmarks for \ourmodel{-on-device} and \ourmodel{-server} alongside relevant sampled comparisons. GSM8K is 8-shot and MATH is 4-shot.
    All results are collected with an internal automated evaluation pipeline.
     }
    \label{fig:math}
\end{figure}

In \autoref{fig:math}, we compare post-training \ourmodel{}'s performance on math benchmarks including GSM8K~\cite{gsm8k} and MATH~\cite{math}. We use $8$-shot chain-of-thought (CoT)~\cite{wei2022chain} prompt for GSM8K and $4$-shot CoT prompt~\cite{lewkowycz2022solving} for MATH. We conduct all evaluations using an internal automated evaluation pipeline. We see that the \ourmodel{-on-device} significantly outperforms \mistral{} and \gemmal{}, even at less than half of their sizes.

\inputhide{tables/math_eval}

\subsection{Summarization feature evaluation}
\label{sec:eval_feature}
The product team specifications for summarizing Emails, Messages, and Notifications necessitated a tailor-made set of guidelines, metrics, and specialized graders to evaluate summarization quality against various open-source, licensed, and proprietary datasets. 

\paragraph{Datasets.}
We sampled abundant payloads carefully for each use case. 
These evaluation datasets emphasize a diverse set of inputs which our product features are likely to face in production, and include a stratified mixture of single and stacked documents of varying content types and lengths. %
We developed a pipeline to build evaluation datasets that simulate real user inputs.

\paragraph{Graders.}
We enlisted a pool of highly-trained, full-time, Apple-employed human graders with specialized writing and comprehension skills to evaluate summarization quality. To qualify for grading projects, each grader must pass a series of eligibility and training steps, which include a required bachelor's degree in a writing-related discipline, customized training sessions, and consistently high performance against internal grading quality benchmarks. 

\paragraph{Grading guidelines.}
During the evaluation task, graders are presented with a specification for the summary, the original input content, and the output summary. Graders assess the summary on each the following sub-dimensions of quality using 3 point scales (``good'', ``neutral'', or ``poor'' ):
\begin{itemize}[itemsep=0pt]
\item[] \textbf{Composition}:  Evaluates the overall readability of the summary considering grammar, punctuation, spelling, and brevity.
\item[] \textbf{Comprehensiveness}: Evaluates how comprehensive the summary is in capturing the essential points or calling out any actions/conclusions for the user.
\item[] \textbf{Groundedness}: Evaluates how grounded the summary is with respect to the original payload. Summaries that are not completely grounded may contain details that are exaggerated, inferred, inaccurate, or hallucinated.
\item[] \textbf{Following instructions}: Evaluates whether the summary meets specific style and formatting requirements. Requirements are tailored to each feature and reflect specific product and design expectations.
\item[] \textbf{Harmfulness}: Evaluates whether the summary contains content that is harmful or unsafe according to Apple's safety taxonomy.
\end{itemize}

A summary is classified as ``poor'' if \textit{any} of the sub-dimensions are ``poor'' according to predefined product specifications. Likewise a summary is ``good'' only if \textit{all} sub-dimensions are good. These classifications are used to compute ``Good/Poor Result Ratio'' metrics defined as the percentage of good/poor summaries out of all summaries.

\begin{figure}[!ht]
    \centering
     \includegraphics[width=1.1\textwidth, alt={Chart with bars showing the good result ratio and poor result ratio for email, messages, and notification use cases.},center]{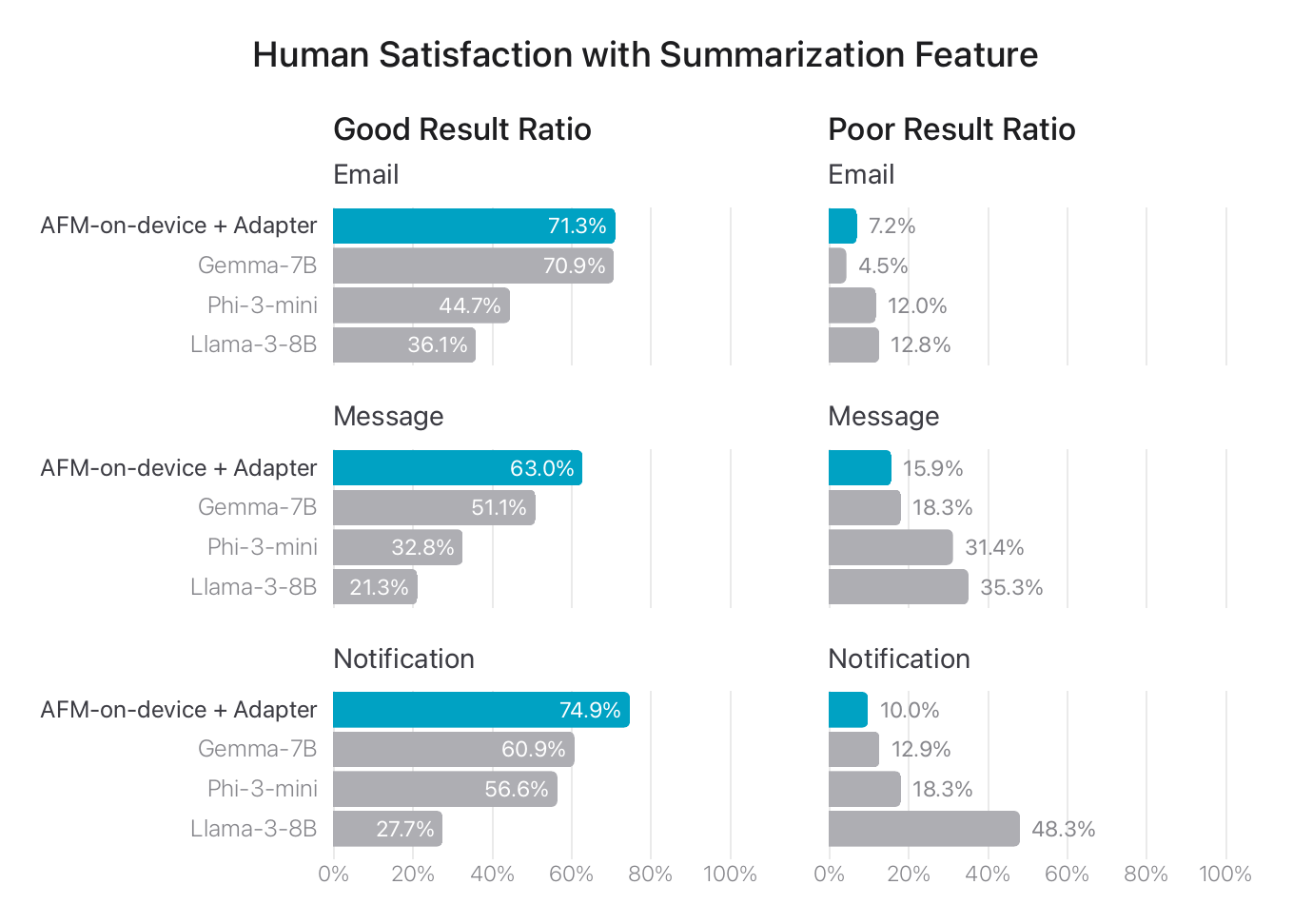}
    \caption{Ratio of ``good'' and ``poor'' responses for three summarization use cases relative to all responses. Summaries are classified as ``good'', ``neutral'', or ``poor'' along five dimensions. A result is classified as ``good'' if \textit{all} of the dimensions are good (higher is better). A result is classified as ``poor'' if \textit{any} of the dimensions are poor (lower is better). Overall, our \ourmodel-on-device adapter{} generates better summaries than comparable models.}
    \label{fig:summ-human-eval}
\end{figure}

\paragraph{Results.}
We ask human graders to evaluate the summarization quality of the \ourmodel-on-device adapter{},  \phimini{}, \llamathrees{}, and \gemmal{}. 
\autoref{fig:summ-human-eval} shows that \ourmodel-on-device-adapter{} overall outperforms the other models.

\inputhide{tables/summarization_domain}

\section{Responsible AI}  %
\label{sSafety}

\subsection{Overview}

Apple Intelligence is developed responsibly and designed with care to empower our users, represent them authentically, and protect their privacy. Of primary importance to our Responsible AI approach is that we are ultimately delivering intelligent, well-defined tools that address specific user needs. Having a clear definition of what a feature is intended to do allows us to better identify any potential safety gaps.

We have developed a safety taxonomy in order to be comprehensive and consistent in the design and evaluation of our generative AI-powered features. This taxonomy builds and extends Apple's extensive experience in using artificial intelligence and machine learning to deliver helpful features to users around the world, and is updated regularly as we develop and test features. Currently, it consists of 12 primary categories comprised of 51 subcategories, including ``Hate Speech, Stereotypes, and Slurs'', ``Discrimination,
Marginalization, and Exclusion'', ``Illegal Activities'', ``Adult Sexual Material'', and ``Graphic Violence.''

The taxonomy serves as a structured way to consider potential issues and risks relative to each specific feature. As new or additional risks are identified, we develop and revise the associated policies that are contextualized to each individual feature, taking into account the specific needs that it serves, the content it produces, and the appropriate mitigations.
They are developed with extensive internal and external input from academics, AI ethicists, trust and safety, and legal experts to better identify and understand the relevant risks, the potential severity of such risks, and the potential disparate impact these risks may have on certain groups.
These policies guide our work in data collection, human annotation, model training, guardrails development, evaluation, and red teaming.

Particularly, the taxonomy is not itself the sole determinant of our policy. For example, content that may fall within the safety taxonomy is not necessarily always blocked, as doing so unilaterally may be in conflict with other aspects of Apple's Responsible AI development principles, such as “respecting how our users choose to use these tools to accomplish their goals.'' Thus, features that operate as tools may be more permissive in the kinds of content they operate over and produce in order to effectively address the user's intent. On the other hand, features that may generate content beyond a user's specified intent may need to be more constrained. Regardless, we strive for some categories of harm to always be treated with special care (such as any content that relates to self harm) while other categories will always be blocked (such as illegal content). 

In addition, our Responsible AI principles are built into every stage of Apple Foundation Models and Apple Intelligence as well as the safety taxonomy, which helps us evaluate risks and formulate policies feature by feature. We include safety-oriented data as part of our fine-tuning of  specific adapters tailored by use case. Furthermore, at the time of inference, we also run guardrail models~\cite{inan2023llama} as pre- and post-processing steps to evaluate potential harm at both the input and output level. Finally, we have mechanisms in place to continuously and proactively improve our AI tools with the help of ongoing user feedback.

\subsection{Pre-Training}

At the pre-training stage, we take several steps to ensure that the values as outlined above are upheld. We follow a strict data policy ensuring that no Apple user data is included, as well as conduct rigorous legal review for each component in the training corpus. Further, we perform safety filtering to reduce potentially harmful content, including NSFW content, profanity, spam, and PII or financial data.

Because pre-training is a step which is shared among various downstream features, our safety mitigations aim to retain general capabilities that allow us to iterate on the taxonomy and policy at a per-feature level, without hurting the helpfulness of these downstream models. We take learnings from prior work to avoid overly aggressive filtering at the pre-training stage, which has potential benefits in safety alignment~\cite{llama2}. Intuitively, the pre-trained model should be aware of content that downstream features and policies may require it to handle -- in some cases with care, or in other cases operating over such content directly.

\subsection{Post-Training}
In the post-training phase, we aim to instill a baseline level of alignment with our Responsible AI principles to avoid necessitating the full complexities of post-training (such as RLHF) in each downstream model that builds on top of the foundation model. In doing so, there are two key considerations:

\begin{enumerate}
    \item We must ensure our models produce output that is helpful to users, while minimizing potential harm.
    \item We must contextualize our safety taxonomy and policies on a feature by feature basis to deliver the best possible user experience.
\end{enumerate}

To balance helpfulness and harmlessness trade-off, our solution is to treat safety alignment as one of the many core post-training tasks that are evaluated and iterated on in tandem, instead of as a separate stage of training. Specifically, we include adversarial data into our SFT and RLHF training corpora that is curated according to our policy and values by partnering closely with trusted vendors. We also incorporate safety tasks and benchmarks into the automatic and human evaluations used during model development.

In total, over 10\% of the training data are adversarial or related to safety or sensitive topics, including single and multi-turn safety category annotations, pairwise and overall preference ratings, and annotator rewrites. This data is either used directly or as seed data for synthetic data generation, as described in \autoref{sec:synthetic_data}.

We do additional work to achieve appropriate safety behavior for each feature beyond baseline alignment. A primary way that we do this is by collecting safety-specific training data and including it when fine-tuning adapters. For instance,
in fine-tuning our summarization adapter we sought to improve aspects such as, improving robustness against malicious questions included within the content to be summarized, and reducing the likelihood that summaries would inadvertently amplify harmful or sensitive content to be summarized.

\subsection{Guarding against malicious code}
Code generation requires special care. Our code benchmarks involve actually executing the generated code to determine both syntactic and semantic correctness. Thus, responsible training of code models involves treating all generated code as unsafe by default -- all code is always executed in a fully locked down environment with no access to the internet or any internal or external services. Specifically, the locked down environment is managed with FireCracker~\cite{agache2020firecracker}, with a FireCracker jailer at the cluster level.

\subsection{Red teaming}
\label{sec:red_teaming}

Red teaming attempts to elicit safety policy violating responses from models, or harmful responses for which no  policy yet exists. These results inform both policy development as well as the focus and content of safety evaluation datasets. These in turn can influence design, engineering, and shipping readiness decisions.   

Red teaming is a fundamentally creative endeavor that requires red teamers to employ combinations of attack vectors to probe known model vulnerabilities, and try to discover new ones. Attack vectors used when engaging with language models include jailbreaks/prompt injections, persuasive techniques~\cite{zeng2024persuade}, and linguistic features known to cause model misbehavior (e.g.\ slang, code-switching, emojis, typos).

We employ both manual and automatic red-teaming~\cite{ganguli2022red} to elicit potentially unknown failure modes of the aligned models. More recent works~\cite{llama2} suggest that automated processes can potentially generate even more diverse prompts than humans, previously seen as the ``gold'' standard for data collection. These automated processes can include using the language models themselves to identify gaps, some of which may be unintuitive or even surprising. Such examples can be used directly as synthetic training or evaluation data and to inform future data collection efforts.

A basic human red teaming task schema is as follows: a red teamer is assigned a safety taxonomy category and attack vector(s). They author an input to the model, using that attack vector, that is intended to elicit a response containing content from that category. If the response does not contain the target content, the red teamer can engage in a fixed number of conversational turns, after which they provide a final harmfulness rating of the model output and list the taxonomy categor(ies) in it, if any. To ensure annotation quality, red teamers also provide an overall confidence score for their ratings. 

In addition to red teaming at the base model level, we also red team specific features. Red teaming projects at the feature level use feature-specific guidelines with attack vectors informed by the feature's safety policy and engineering concerns. These projects can provide in-depth probing of known risks for that particular feature and also adversarially probe for unknown vulnerabilities.

Our red teaming projects are run using internal and external crowds. To ensure responsible data collection, due to the sensitive nature of red teaming we: 1) make red teaming completely voluntary; 2) impose a strict time limit on how much each red teamer spends on the tasks per week; 3) provide health and well-being resources available around the clock; and 4) maintain an open line of communication with internal red teamers via weekly office hours and a Slack channel for them to communicate any concerns that arise.

\subsection{Evaluation}
\label{sec:safety-eval}

As mentioned in previous sections, safety is one of the many axes iterated on during foundation model development, and therefore undergoes the same automatic and human evaluation cycles during post-training.

\paragraph{Safety evaluation set}

To reduce noise, cost, and turn-around time during human evaluations, we must ensure that our safety evaluation sets are clean, yet challenging and comprehensive. To that end, we filter out ``easy" prompts which consistently yield low harmfulness responses across different versions of the model, and employ an embedding-based analysis to improve our evaluation prompt set coverage. Overall, we curate a set of over a thousand adversarial prompts to test \ourmodel{}'s performance on harmful content, sensitive topics, and factuality according to our safety policy.

\inputhide{tables/human_eval_safety}

\begin{figure}[!ht]
    \centering
     \includegraphics[width=1.1\textwidth, alt={Chart with bars showing the performance of various models for output harmfulness. On-device models are on the left and server models are on the right.},center]{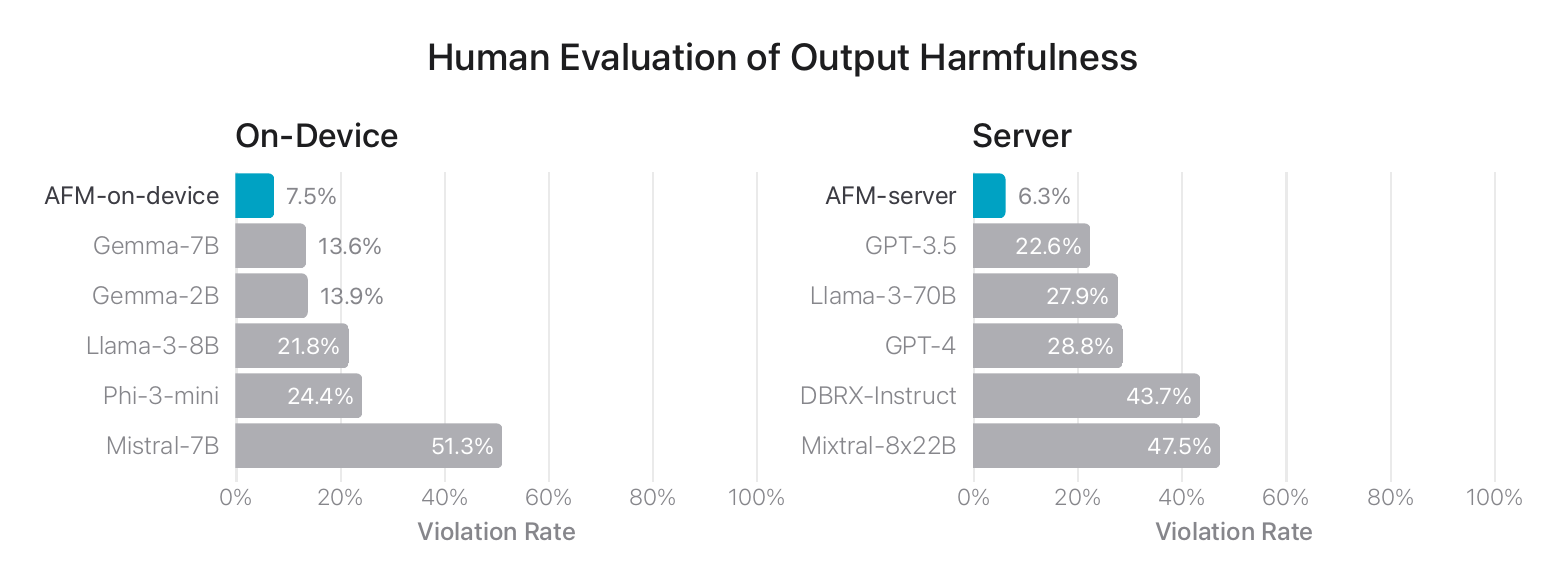}
    \caption{Fraction of violating responses for harmful content, sensitive topics, and factuality (lower is better). Our models are robust when faced with adversarial prompts.
    }
    \label{fig:human-eval-harmfulness}
\end{figure}

\begin{figure}[!ht]
    \centering
     \includegraphics[width=1.0\textwidth, alt={Chart showing which models win, tie, or loose in a direct comparison for safety prompts. On-device comparisons are at the top and server comparisons are at the bottom.},center]{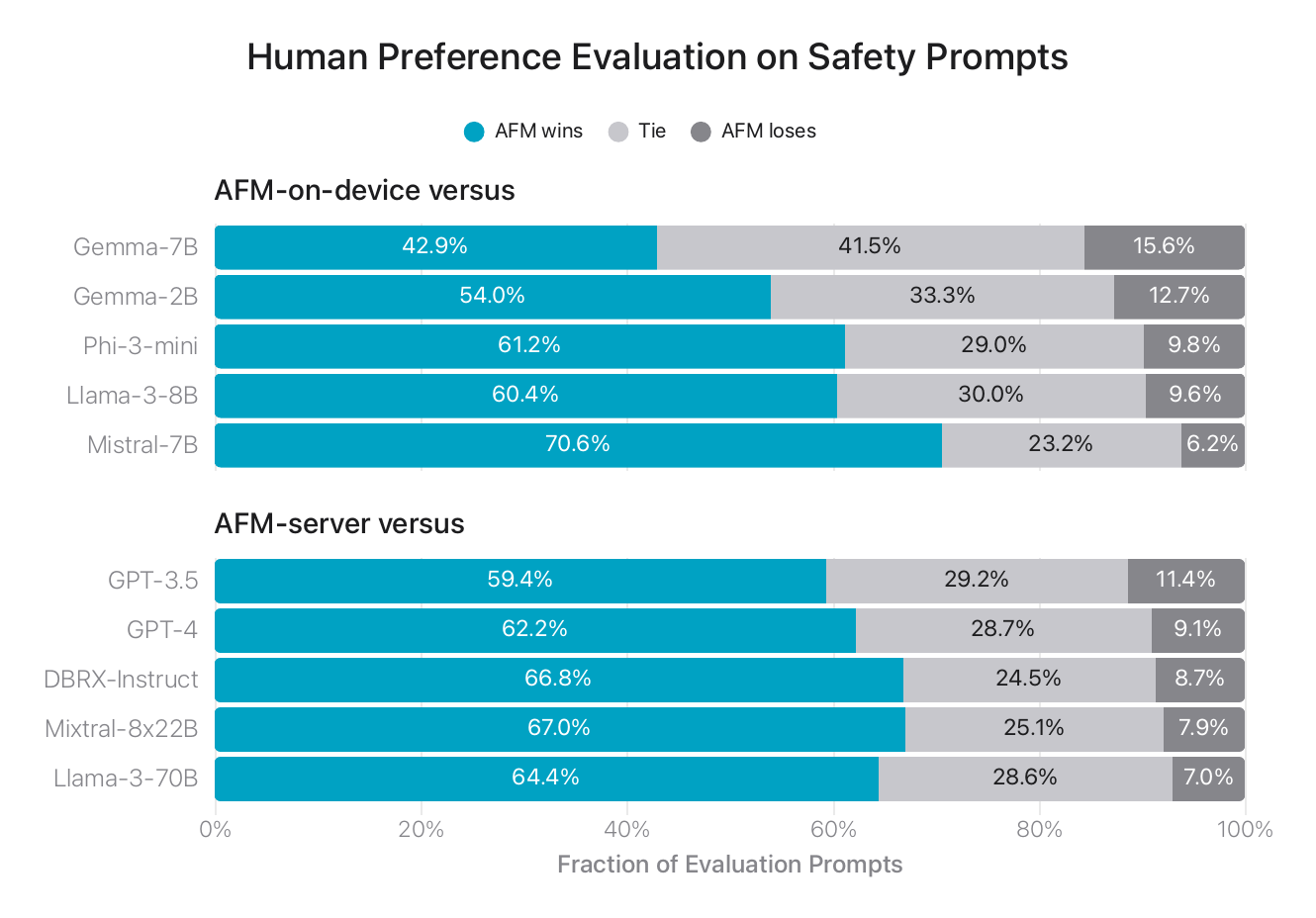}
    \caption{Fraction of preferred responses in side-by-side evaluation of Apple's foundation model against comparable models on safety prompts. Human graders found our responses safer and more helpful.
    }
    \label{fig:human-eval-sxs}
\end{figure}

\paragraph{Safety evaluation results} 
\autoref{fig:human-eval-harmfulness} summarizes the violation rates of different models evaluated by human graders on this safety evaluation set. Lower is better.
Both \ourmodel{-on-device} and \ourmodel{-server} are robust to adversarial prompts,
achieving violation rates significantly lower than open-source and commercial models.
In addition, we report side-by-side human preference on our safety evaluation prompts in 
 \autoref{fig:human-eval-sxs}. \ourmodel{} models are preferred by human graders as safe and helpful responses over competitor models.

\section{Conclusion}
In this report we introduced the foundation language models that power Apple Intelligence features, \ourmodel{-on-device} and \ourmodel{-server}. 
The models are designed to be fast and run efficiently on iPhone, iPad, and Mac as well as on Apple silicon servers via Private Cloud Compute.
They are trained to be highly capable in tasks like language understanding, instruction following, reasoning, writing, and tool use.
We have developed an innovative model architecture to specialize these models for our users' most common tasks.
On top of the foundation models, feature-specific adapters are fine-tuned to provide high-quality user experiences such as summarization of emails, messages, and notifications.
Our models have been created with the purpose of helping users do everyday activities across their Apple products, grounded in Apple's core values, and rooted in our Responsible AI principles at every stage.
These foundation models are at the heart of Apple Intelligence, the personal intelligence system built by Apple to continue empowering our users and enriching their lives.

\bibliography{main}
\clearpage
\section*{Contributors}
Within each section, contributors are listed in alphabetical order by first name.
\subsection*{Foundation Models}
Andy Narayanan,
Aonan Zhang,
Bowen Zhang,
Chen Chen,
Chong Wang (inference efficiency lead),
Chung-Cheng Chiu,
David Qiu,
Deepak Gopinath,
Dian Ang Yap,
Dong Yin,
Feng Nan,
Floris Weers,
Guoli Yin,
Haoshuo Huang,
Jianyu Wang,
Jiarui Lu,
John Peebles,
Ke Ye,
Mark Lee,
Nan Du,
Qibin Chen,
Quentin Keunebroek,
Ruoming Pang (overall lead),
Sam Wiseman,
Syd Evans,
Tao Lei,
Tom Gunter (pre-train lead),
Vivek Rathod,
Xiang Kong,
Xianzhi Du,
Yanghao Li,
Yongqiang Wang,
Yuan Gao,
Zaid Ahmed,
Zhaoyang Xu,
Zhiyun Lu,
Zirui Wang (post-train lead)

\subsection*{Data, Evaluation, and Responsible AI}
Al Rashid,
Albin Madappally Jose,
Alec Doane,
Alfredo Bencomo,
Allison Vanderby,
Andrew Hansen,
Ankur Jain,
Anupama Mann Anupama,
Areeba Kamal,
Bugu Wu,
Carolina Brum,
Charlie Maalouf,
Chinguun Erdenebileg,
Chris Dulhanty,
Daniel Parilla,
Dominik Moritz,
Doug Kang,
Eduardo Jimenez,
Evan Ladd,
Fangping Shi,
Felix Bai,
Frank Chu,
Fred Hohman,
Hadas Kotek,
Hannah Gillis Coleman,
Jane Li,
Jeffrey Bigham,
Jeffery Cao,
Jeff Lai,
Jessica Cheung,
Jiulong Shan,
Joe Zhou,
John Li,
Jun Qin,
Karanjeet Singh,
Karla Vega,
Ke Ye,
Kelvin Zou,
Laura Heckman,
Lauren Gardiner,
Margit Bowler,
Mark Lee,
Maria Cordell,
Meng Cao,
Nicole Hay,
Nilesh Shahdadpuri,
Otto Godwin,
Pranay Dighe,
Pushyami Rachapudi,
Ramsey Tantawi,
Roman Frigg,
Sam Davarnia,
Sanskruti Shah,
Saptarshi Guha,
Sasha Sirovica,
Shen Ma,
Shuang Ma,
Simon Wang,
Sulgi Kim,
Suma Jayaram,
Vaishaal Shankar,
Varsha Paidi,
Vivek Kumar,
Xiang Kong,
Xin Wang,
Xin Zheng,
Walker Cheng,
Yael Shrager,
Yang Ye,
Yasu Tanaka,
Yihao Guo,
Yunsong Meng,
Zhao Tang Luo,
Zhi Ouyang,
Zhiyun Lu

\subsection*{Adapters, Optimizations, and Summarization}
Alp Aygar,
Alvin Wan,
Andrew Walkingshaw,
Andy Narayanan,
Antonie Lin,
Arsalan Farooq,
Brent Ramerth,
Chong Wang,
Colorado Reed,
Chris Bartels,
Chris Chaney,
David Riazati,
Eric Liang Yang,
Erin Feldman,
Gabriel Hochstrasser,
Guillaume Seguin,
Guoli Yin,
Irina Belousova,
Jianyu Wang,
Joris Pelemans,
Karen Yang,
Keivan Alizadeh Vahid,
Liangliang Cao,
Mahyar Najibi ,
Marco Zuliani,
Max Horton,
Minsik Cho,
Nikhil Bhendawade,
Patrick Dong,
Piotr Maj,
Pulkit Agrawal,
Qi Shan,
Qibin Chen,
Qichen Fu,
Regan Poston,
Sam Xu,
Shuangning Liu,
Sushma Rao,
Tashweena Heeramun,
Thomas Merth,
Uday Rayala,
Victor Cui,
Vivek Rangarajan Sridhar,
Vivek Rathod,
Wencong Zhang,
Wenqi Zhang,
Wentao Wu,
Xiang Kong,
Xingyu Zhou,
Xinwen Liu,
Yang Zhao,
Yin Xia,
Zhile Ren,
Zhongzheng Ren

\newpage
\appendix
\section*{Appendix}
\section{Core pre-training recipe ablation}\label{sec:core_pretrain_recipe_ablation}
We compare our chosen settings for `core' pre-training from \autoref{sec:core_pretraining} (optimizer, scaling-law-predicted batch-size, weight-decay, etc.) to a baseline based on~\cite{wortsman2023}. In particular, the baseline uses AdamW with a standard hyperparameter configuration of $\beta_1=0.9$, $\beta_2=0.95$, $\epsilon=1\mathrm{e}{-15}$, and a decoupled weight decay of $1\mathrm{e}{-4}$, decaying the learning rate to $0.0001$ of peak, with a batch size of 1024 sequences. Otherwise both recipes are identical. Training covers 3.1T tokens using the \ourmodel{-on-device} architecture but with a different data mixture to that used by the official \ourmodel{} training runs.

\begin{table}[!ht]
    \centering
    \begin{tabular}{@{}lrr@{}}
        \toprule
        \textbf{Task} & \textbf{Baseline (acc)} & \textbf{\ourmodel{} (acc)} \\
        \midrule
        \textbf{arc\_challenge} & 41.9 & 44.6 \\ 
        \textbf{arc\_easy} & 75.6 & 76.1 \\ 
        \textbf{hellaswag} & 54.3 & 55.0 \\ 
        \textbf{lambada} & 69.3 & 68.9 \\ 
        \textbf{piqa} & 78.3 & 78.4 \\ 
        \textbf{sciq} & 94.5 & 94.7 \\ 
        \textbf{winogrande} & 67.3 & 66.9 \\ 
        \textbf{triviaqa (1-shot)} & 40.5 & 41.0 \\ 
        \textbf{webqs (1-shot)} & 20.6 & 20.6 \\ \midrule
        \textbf{CoreEN average} & 60.2 & 60.7 \\ \midrule
        \textbf{GSM8K (8-shot CoT)} & 16.6 & 18.9 \\ 
        \textbf{MMLU (5-shot)} & 45.4 & 45.5 \\
        \bottomrule
    \end{tabular}
    \caption{Core pre-training recipe ablation few-shot results. Unless otherwise noted, we use 0-shot prompts. We note that \ourmodel{}'s recipe allows for slight improvements across the majority of tasks, although the difference is typically very small. The data mixture differs from the official \ourmodel{} runs.}
    \label{table:pre-trainingablation}
\end{table}

In \autoref{table:pre-trainingablation}, \ourmodel{}'s recipe demonstrates a slight improvement over the baseline. This likely indicates that the most important recipe settings are already well-enough configured by the baseline for this model size and training budget.

\section{Ablations on pruning and distillation}\label{sec:abl_prune_distill}
Here we detail the evaluation results of using structural pruning and distillation separately and show they can be combined together to get the best performance. 

\autoref{tab:v3-prune-distill} shows the ablation results of training 3B models using an early version of our pre-training data mixture. 
As shown in the table, both pruning and distillation methods can outperform a baseline model trained from scratch. For example, pruning and distillation achieve a MMLU score of 42.9\% and 44.9\% respectively, whereas a baseline using 50\% more steps gets 34.6\%.
It is also interesting that pruning achieves a higher score on the CoreEn benchmark, while distillation is better on MMLU.
Finally, when combining these two methods together, we observe further improvements on MMLU and GSM8k by a large margin, getting better or on par results compared to the baseline trained using 5$\times$ more computation.

\begin{table}[h]
    \centering
    \begin{tabular}{@{}l|rrrr|r@{}}
        \toprule
        \textbf{Metric/Method} & \textbf{Baseline} & \textbf{Prune} & \textbf{Distill} & \textbf{Both} & \textbf{Baseline} \\
        \midrule
        \textbf{Training cost} & 1.5$\times$ & 1$\times$ & 1$\times$ & 1$\times$ & 5$\times$ \\ 
        \midrule
        \textbf{MMLU (5-shot)} & 34.6 & 42.9 & 44.9 & \bftab 49.3  & 45.4 \\
        \textbf{GSM8K (8-shot CoT)} & 12.7 & 13.5 & 11.0 & \bftab 16.8 & \bftab 16.9 \\
        \textbf{CoreEN Average} & 59.8 & \bftab 61.0 & 58.1 & 59.7 & 60.3 \\
        \bottomrule
    \end{tabular}
    \caption{Ablation results of pruning and distillation methods. The training data is an early version that differs from the official AFM runs.}
    \label{tab:v3-prune-distill}
\end{table}

\section{Pre-training stage-breakdown evaluations}\label{sec:pre-train_stage_breakdown_eval}

We present few-shot evaluation results after core, continued, and long-context pre-training stages, for a subset of evaluation metrics that we find to be low-variance, diverse, and correlated with downstream evaluation after post-training. These metrics are derived using an internal harness and set of benchmark formulations, which are not optimized for absolute performance (e.g. we do not apply length normalization, and use more difficult test splits where available---for TriviaQA as one example). They are therefore not suitable for comparison with other published results.

\begin{table}[!ht]
    \centering
    \begin{tabular}{@{}lrrr@{}}
        \toprule
        \textbf{\ourmodel{-on-device}} & \textbf{Core} & \textbf{Continued} & \textbf{Context lengthened} \\ \midrule
        \textbf{ARC\_C} & 43.17 & 47.53 & 45.39 \\ 
        \textbf{ARC\_E} & 74.87 & 78.62 & 78.37 \\ 
        \textbf{HellaSwag} & 54.70 & 55.50 & 55.24 \\ 
        \textbf{LAMBADA} & 73.51 & 70.13 & 69.90 \\ 
        \textbf{PIQA} & 77.37 & 78.67 & 78.40 \\ 
        \textbf{SciQ} & 94.90 & 95.80 & 95.70 \\ 
        \textbf{WinoGrande} & 65.82 & 67.32 & 67.01 \\ 
        \textbf{TriviaQA (1 shot)} & 42.46 & 39.13 & 38.11 \\ 
        \textbf{WebQS (1 shot)} & 19.24 & 18.06 & 17.22 \\ 
        \midrule
        \textbf{CoreEN average} & 60.67 & 61.20 & 60.59 \\ 
        \midrule
        \textbf{MMLU (5 shot)} & 57.00 & 61.35 & 60.64 \\ 
        \textbf{GSM8K (8 shot CoT)} & 27.45 & 42.53 & 40.00 \\ 
        \textbf{MATH (4 shot CoT)} & 8.31 & 16.97 & 15.48 \\ 
        \textbf{HumanEval-Py pass@1} & 16.48 & 27.38 & 30.84 \\ 
        \textbf{MultiPLE-Swift pass@1} & 8.88 & 19.24 & 18.06 \\ 
        \bottomrule
    \end{tabular}
    \caption{Pre-training evaluation for \ourmodel{-on-device} with an internal harness. Unless otherwise noted, we use 0-shot prompts. TriviaQA evaluation is on the larger and more challenging ``Web'' split.}
    \label{table:3bpre-traininternal}
\end{table}

\begin{table}[!ht]
    \centering
    \begin{tabular}{@{}lrrr@{}}
        \toprule
        \textbf{\ourmodel{-server}} & \textbf{Core} & \textbf{Continued} & \textbf{Context lengthened} \\ \midrule
        \textbf{ARC\_C} & 58.28 & 58.87 & 57.94 \\ 
        \textbf{ARC\_E} & 85.61 & 85.44 & 85.06 \\ 
        \textbf{HellaSwag} & 64.17 & 64.53 & 64.37 \\ 
        \textbf{LAMBADA} & 78.38 & 77.59 & 77.82 \\ 
        \textbf{PIQA} & 82.37 & 81.99 & 81.88 \\ 
        \textbf{SciQ} & 96.60 & 97.10 & 97.00 \\ 
        \textbf{WinoGrande} & 80.51 & 79.16 & 79.08 \\ 
        \textbf{TriviaQA (1 shot)} & 54.33 & 53.57 & 53.42 \\ 
        \textbf{WebQS (1 shot)} & 29.97 & 27.66 & 27.41 \\ 
        \midrule
        \textbf{CoreEN average} & 70.02 & 69.55 & 69.33 \\ 
        \midrule
        \textbf{MMLU (5 shot)} & 74.00 & 75.24 & 74.80 \\ 
        \textbf{GSM8K (8 shot CoT)} & 75.44 & 74.83 & 75.51 \\ 
        \textbf{MATH (4 shot CoT)} & 32.24 & 36.48 & 35.77 \\ 
        \textbf{HumanEval-Py} & 33.23 & 40.77 & 39.55 \\ 
        \textbf{MultiPLE-Swift} & 30.15 & 37.70 & 38.11 \\
        \bottomrule
    \end{tabular}
    \caption{Pre-training evaluation for \ourmodel{-server} with an internal harness. Unless otherwise noted, we use 0-shot prompts. TriviaQA evaluation is on the larger and more challenging ``Web" split.}
    \label{table:30bpre-traininternal}
\end{table}

In \autoref{table:3bpre-traininternal} and \ref{table:30bpre-traininternal} we present internal benchmarks after all three stages of pre-training. As expected, continued pre-training acts to improve math and particularly code model capabilities, whilst subtly improving a few other benchmarks. The context-lengthening stage leaves the majority of these benchmarks on-par, with changes (positive and negative) typically within the range of what we consider to be evaluation noise.

\section{Long-context evaluation}\label{sec:ruler_long_context}

Although the focus for this version of \ourmodel{} was not to support context lengths longer than 8k, in \autoref{table:ruler} we use the RULER~\cite{hsieh2024rulerwhatsrealcontext} benchmark to evaluate \ourmodel{-server} at 4k to 32k context lengths. We note that the model is capable of performing perfectly up to a sequence length of $\geq\mathrm{32k}$ when tested against straightforward retrieval-like tasks, e.g., needle-in-the-haystack (NIAH). It is clear, however, that the model performance gradually suffers with an increasing context length on RULER, a more complex evaluation benchmark than NIAH, suggesting that the real context length for \ourmodel{-server}, for tasks beyond retrieval, is currently at most 24k.

\begin{table}[!ht]
    \centering
    \begin{tabular}{@{}lr@{}}
        \toprule
        \textbf{AFM-server} & \textbf{Average acc} \\ \midrule
        \textbf{Ctx @ 4096} & 91.7 \\ 
        \textbf{Ctx @ 8192} & 87.7 \\ 
        \textbf{Ctx @ 16384} & 84.1 \\ 
        \textbf{Ctx @ 20480} & 79.1 \\ 
        \textbf{Ctx @ 24576} & 75.8 \\ 
        \textbf{Ctx @ 32768} & 43.3 \\ 
        \bottomrule
    \end{tabular}
    \caption{RULER~\cite{hsieh2024rulerwhatsrealcontext} average evaluation results, averaged over 13 synthetic long-context tasks using 500 examples per task.}
    \label{table:ruler}
\end{table}

\section{Technical details for RLHF}\label{sec:appendix_hlhf}

\subsection{Reward modeling}\label{sec:appendix_reward_model}
The human preference data that we use in reward model training has the following format:
\begin{itemize}
    \item $x$: the prompt;
    \item $y_c$: the chosen (preferred) response;
    \item $y_r$: the rejected response;
    \item $\ell$: the level of the human preference;
    \item $z^{\text{if}}_c$ and $z^{\text{if}}_r$: the instruction-following property of the two responses;
    \item $z^{\text{verb}}_c$ and $z^{\text{verb}}_r$: the verbosity of the two responses;
    \item $z^{\text{truth}}_c$ and $z^{\text{truth}}_r$: the truthfulness of the two responses;
    \item $z^{\text{harm}}_c$ and $z^{\text{harm}}_r$: the harmlessness of the two responses.
\end{itemize}
In our reward modeling, the preference level $\ell$ takes $4$ possible values, indicating that the chosen response is negligibly better, slightly better, better, or significantly better than the rejected response. As for the single sided gradings, each label, e.g., $z^{\text{if}}_c$, takes $3$ possible values. For instruction following, truthfulness, and harmlessness, the $3$ values correspond to the cases where the response has major issue, minor issue, or no issue. For verbosity, the $3$ values correspond to the cases where the response is too verbose, too short, or just right.

We use a multi-head architecture for the reward model. More specifically, we take a decoder-only transformer and obtain the last-layer embedding of the last non-padding token. We attach one linear and four MLP heads to the embedding. Denote the model parameters by $\phi$ and the input prompt-response pair by $(x, y)$. The linear head outputs the preference reward $r_\phi(x, y)\in\mathbb{R}$. The four MLP heads are classification heads representing the instruction-following, verbosity, truthfulness, and harmlessness property of the response. We denote the output logits of the $4$ classification heads by $u_\phi^{\text{if}}$, $u_\phi^{\text{verb}}$, $u_\phi^{\text{truth}}$, $u_\phi^{\text{harm}}$, respectively.

\paragraph{Soft label loss.} We train the preference reward $r_\phi(x, y)$ based on Bradley-Terry-Luce (BTL) model~\cite{bradley1952rank}. Recall that in BTL model, the probability that $y_c$ is preferred over $y_r$ is modeled as $\sigma(r_\phi(x, y_c) - r_\phi(x, y_r))$, where $\sigma$ is the sigmoid function. Intuitively, this probability should be larger if the preferred response $y_c$ is annotated as significantly better than the rejected response $y_r$, and smaller if $y_c$ is only negligibly better than $y_r$. We incorporate this information using the preference level $\ell$. More specifically, for each preference level $\ell$, we design a \emph{target} preference probability $p_\ell$. Then we use a \emph{soft label} loss as follows:
\begin{equation}\label{eq:soft_label}
\begin{aligned}
L_{\text{ranking}}(\phi) = &-p_\ell \log(\sigma(r_\phi(x, y_c) - r_\phi(x, y_r))  \\
&- (1-p_\ell) \log(\sigma(r_\phi(x, y_r) - r_\phi(x, y_c)).
\end{aligned}
\end{equation}
The target level $p_\ell$ is a hyperparameter in our algorithm and should take larger value if the preference level is higher. In our experiments, we choose $p_\ell=0.95,~0.85,~0.75,~0.65$ for significantly better, better, slightly better, and negligibly better, respectively.

\paragraph{Single-sided grading as regularization.} We also leverage the single-sided gradings as regularization terms in our reward model. The intuition is that with these gradings as regularization terms, we can learn a better embedding to capture human preferences. The regularization loss is
\begin{equation}\label{eq:regularization}
    \begin{aligned}
    L_{\text{regu}}(\phi) =& \sum_{\text{grade}\in{\text{if,verb,truth,harm}}} \Big( \mathsf{cross\_entropy}(u_\phi^\text{grade}(x, y_c), z^{\text{grade}}_c)  \\
    &+ \mathsf{cross\_entropy}(u_\phi^\text{grade}(x, y_r), z^{\text{grade}}_r) \Big).
    \end{aligned}
\end{equation}
Overall, the reward model training loss that we use is
\begin{align}\label{eq:reward_model_loss}
L_{\text{ranking}}(\phi) + \lambda L_{\text{regu}}(\phi).
\end{align}

\subsection{Online RL algorithm}\label{sec:appendix_rl}
In this section, we present more details of our online RLHF algorithm, MDLOO.

\paragraph{Leave-One-Out (LOO) estimator of the advantage.}
In each iteration of the algorithm, we have a data collection stage and a policy updating stage. Let $\theta_k$ be the model parameter at the beginning of the $k$-th iteration. We sample a batch of $n$ prompts from our prompt set, and for each prompt, we sample $K$ responses according to the policy $\pi_{\theta_k}$, and thus collecting a total of $nK$ data points in each iteration. Let $x$ be a prompt and $y_i$ be one of the responses. Since we consider the bandit setting, by definition, the \emph{advantage} of $(x, y_i)$ is 
\begin{align}\label{eq:adv_def}
A_k(x, y_i) = R(x, y_i) - \mathbb{E}_{y\sim \pi_{\theta_k}(\cdot | x)}[R(x, y)].
\end{align}
We use the leave-one-out (LOO) method~\cite{loo:kool2019buy} to estimate $A_k(x, y_i)$. Namely, we estimate the mean reward given the prompt $x$ with the other $K-1$ responses, i.e.,
\begin{align}\label{eq:adv_est}
\widehat{A}_k(x, y_i) = R(x, y_i) - \frac{1}{K-1}\sum_{j\neq i} R(x, y_j).
\end{align}
As shown in recent works~\cite{ahmadian2024back}, this advantage estimation is beneficial for RLHF. Empirically, we find that using LOO estimator leads to more stable training and better results compared to directly using the reward as the advantage estimation or using the difference between the reward and a running average baseline~\cite{williams1992simple}.

\paragraph{Mirror descent policy optimization (MDPO).} Our policy optimization approach belongs to a widely used class of trust-region policy optimization algorithms~\cite{schulman2015trust}. The basic idea in these algorithms is that in each policy iteration, we apply a regularization method to prevent the policy from changing too much in an iteration. The regularization can be achieved by adding KL regularization~\cite{mdpo:tomar2020mirror,abbasi2019politex,lazic2021improved} and using clipping for the probability ratio such as in PPO~\cite{ppo:schulman2017proximal}. In this work, we use KL regularization as in Mirror Descent Policy Optimization (MDPO)~\cite{mdpo:tomar2020mirror}.

In particular, in the $k$-th iteration, with the data (prompts along with the $K$ responses sampled according to $\pi_{\theta_k}$ for each prompt), we aim to optimize the following regularized advantage maximization problem:
\begin{align}\label{eq:rl_iter_obj}
\max_{\theta} \Psi(\theta):= \mathbb{E}_{x\sim \mathcal{D}}\left[ \mathbb{E}_{y\sim\pi_{\theta_k}(\cdot | x)}[A_k(x, y)] - \gamma D_{\text{KL}}( \pi_{\theta}(\cdot|x) || \pi_{\theta_k}(\cdot|x) )\right].
\end{align}
Note that here the KL regularization term is different from the one in Eq.~\eqref{eq:rl_objective}. The KL regularization in Eq.~\eqref{eq:rl_objective} is between the policy model and the reference model; whereas the KL regularization term in Eq.~\eqref{eq:rl_iter_obj} is between the policy model and the policy at the beginning of the $k$-th iteration. Then we can obtain the gradient of $\Psi(\theta)$ as
\begin{equation}\label{eq:rl_iter_obj_grad}
    \begin{aligned}
        \nabla \Psi(\theta) = &\mathbb{E}_{x \sim \mathcal{D}, y \sim \pi_{\theta_k}(\cdot | x)} \left[\frac{\pi_{\theta}(y|x)}{\pi_{\theta_k}(y|x)} A_k(x, y) \nabla \log \pi_{\theta}(y|x)\right] \\
        & - \gamma \mathbb{E}_{x \sim \mathcal{D}}\left[\nabla D_{\text{KL}}\pi_{\theta}(\cdot | x)||\pi_{\theta_k}(\cdot|x)\right].
    \end{aligned}
\end{equation}
The MDLOO algorithm can be derived by replacing the expectations in Eq.~\eqref{eq:rl_iter_obj_grad} with the $nK$ samples collected with $\pi_{\theta_k}$, and the advantage $A_k(x, y)$ with the LOO estimator $\widehat{A}_k(x, y)$ in Eq.~\eqref{eq:adv_est}. Empirically, we find that MDLOO works better than the popular PPO~\cite{ppo:schulman2017proximal} algorithm in our setting.

\section{Accuracy-recovery adapters ablation}\label{sec:quant_ablation}
In this section, we present the evaluation results on unquantized, quantized, and accuracy-recovered models. As shown in~\autoref{table:qlora}, the quantized models have huge quality drops in both pre-train and post-train metrics. By using accuracy-recovery LoRA adapters with only rank 16, Alpaca win rate can be improved by 7-18\%, GMS8K accuracy is boosted by 5-10\%.
The recovered models perform much closer to the original unquantized model while achieving significant reductions on the model size.
More interestingly, we observe that when the quantization scheme becomes more aggressive (from 3.7 to 3.5 bpw), the adapters also recover more quality back.

\begin{table}[!ht]
    \centering
    \small
    \resizebox{\textwidth}{!}{
    \begin{tabular}{@{}llrrr@{}}
    \toprule
    \bftab BPW & \bftab Models & \bftab IFEval Instruction-Level & \bftab AlpacaEval 2.0 LC & \bftab GSM8K (8-shot CoT) \\ \midrule
    \bftab 16 & \bftab \ourmodel{-on-device} & 100.0\% & 100.0\% & 100.0\% \\ \midrule
    \bftab \multirow{2}{*}{3.5} & \bftab quantized & 98.4\% & 76.7\% & 82.2\% \\
    & \bftab Acc.-recovered (rank 16) & 98.8\% & 94.7\% & 92.1\% \\ \midrule
    \bftab \multirow{2}{*}{3.7} & \bftab quantized & 97.9\% & 87.3\% & 91.3\% \\
    & \bftab Acc.-recovered (rank 16) & 100.6\% & 94.8\% & 96.0\% \\ \bottomrule
    \end{tabular}
    }
    \caption{Evaluation results for quantized and accuracy-recovered models. Numbers are normalized to the unquantized version.}
    \label{table:qlora}
\end{table}

\end{document}